\documentclass{article}

\usepackage{microtype}
\usepackage{graphicx}
\usepackage{subcaption}
\usepackage{booktabs} % for professional tables

\usepackage{hyperref}

\usepackage[utf8]{inputenc}
\usepackage{booktabs} % 必须：用于生成高质量的三线表
\usepackage{multirow} % 必须：用于合并单元格
\usepackage{graphicx} % 必须：用于调整表格宽度
\usepackage{color}
\usepackage{makecell} % 用于表头换行
\usepackage[table]{xcolor}
\newcommand{\gain}[1]{\textcolor[RGB]{0,200,0}{\scriptsize (#1)}}
\usepackage{pifont} % 提供打钩和打叉符号
\usepackage{amssymb} % 备选：提供 \checkmark

\usepackage[preprint]{icml2026}

\usepackage{amsmath}
\usepackage{amssymb}
\usepackage{mathtools}
\usepackage{amsthm}

\usepackage[capitalize,noabbrev]{cleveref}

\theoremstyle{plain}

\theoremstyle{definition}

\theoremstyle{remark}

\usepackage[textsize=tiny]{todonotes}

\icmltitlerunning{Dual-Memory Augmentation for Long-Horizon Web Agents via Trajectory Summarization and Insight Retrieval}

\begin{document}

\twocolumn[
  \icmltitle{M$^2$: Dual-Memory Augmentation for Long-Horizon Web Agents via \\ Trajectory Summarization and Insight Retrieval}



  \icmlsetsymbol{equal}{*}

  \begin{icmlauthorlist}
    \icmlauthor{Dawei Yan}{yyy,comp}
    \icmlauthor{Haokui Zhang*}{yyy}
    \icmlauthor{Guangda Huzhang}{comp}
    \icmlauthor{Yang Li}{comp}
    \icmlauthor{Yibo Wang}{comp}
    \icmlauthor{Qing-Guo Chen}{comp}
    \icmlauthor{Zhao Xu}{comp}
    \icmlauthor{Weihua Luo}{comp}
    \icmlauthor{Ying Li}{yyy}
    \icmlauthor{Wei Dong}{sch}
    \icmlauthor{Chunhua Shen}{ccc}
  \end{icmlauthorlist}

  \icmlaffiliation{yyy}{Northwestern Polytechnical University, Xi'an, China}
  \icmlaffiliation{comp}{AI Business, Alibaba Group, Hangzhou, China}
  \icmlaffiliation{sch}{Xi'an University of Architecture and Technology, Xi'an, China}
  \icmlaffiliation{ccc}{Zhejiang University, Hangzhou, China}

  \icmlcorrespondingauthor{Haokui Zhang}{hkzhang@nwpu.edu.cn}

  \icmlkeywords{Machine Learning, ICML}

  \vskip 0.3in
]



\printAffiliationsAndNotice{}  % no special notice (required even if empty)

\begin{abstract}
Multimodal Large Language Models (MLLMs) based agents have demonstrated remarkable potential in autonomous web navigation. However, handling long-horizon tasks remains a critical bottleneck. Prevailing strategies often rely heavily on extensive data collection and model training, yet still struggle with high computational costs and insufficient reasoning capabilities when facing complex, long-horizon scenarios. To address this, we propose M$^2$, a training-free, memory-augmented framework designed to optimize context efficiency and decision-making robustness. Our approach incorporates a dual-tier memory mechanism that synergizes Dynamic Trajectory Summarization (Internal Memory) to compress verbose interaction history into concise state updates, and Insight Retrieval Augmentation (External Memory) to guide the agent with actionable guidelines retrieved from an offline insight bank. Extensive evaluations across WebVoyager and OnlineMind2Web demonstrate that M$^2$ consistently surpasses baselines, yielding up to a 19.6\% success rate increase and 58.7\% token reduction for Qwen3-VL-32B, while proprietary models like Claude achieve accuracy gains up to 12.5\% alongside significantly lower computational overhead.

\end{abstract}

\begin{figure}[htbp]
    \centering
    % --- 左侧 Minipage：放置表格 ---
    % [c] 表示内容垂直居中对齐。宽度设为当前栏宽的 58%
    \begin{minipage}[c]{0.45\linewidth}
        \centering
        \newcommand{\vmark}{\ding{51}} % 打钩
        \newcommand{\xmark}{\ding{55}} % 打叉
        \captionof{table}{Comparison with SOTA methods on WebVoyager dataset. $\star$: Official metrics from the original release; note that some tasks within this benchmark are now infeasible due to website updates.}
        \label{tab:sota_sidebyside}
        {\scriptsize
        \begin{tabular}{@{}lccc@{}}
        \toprule
        \textbf{Model} & \textbf{SFT} & \textbf{RL} & \textbf{Acc.} \\
        \midrule
        UI-TARS-1.5 & \vmark & \vmark & 84.8\rlap{\(^\star\)} \\
        OpenAI CUA  & \xmark & \vmark & 87.0\rlap{\(^\star\)} \\
        \midrule
        M$^2$ (Ours) & \xmark & \xmark & 86.0 \\
        \bottomrule
    \end{tabular}}
    \end{minipage}% 注意这里不要留空行，否则会换行
    \hfill % 在两个 minipage 之间加入弹性空白，
    \begin{minipage}[c]{0.5\linewidth}
        \centering
        \includegraphics[width=\linewidth, keepaspectratio]{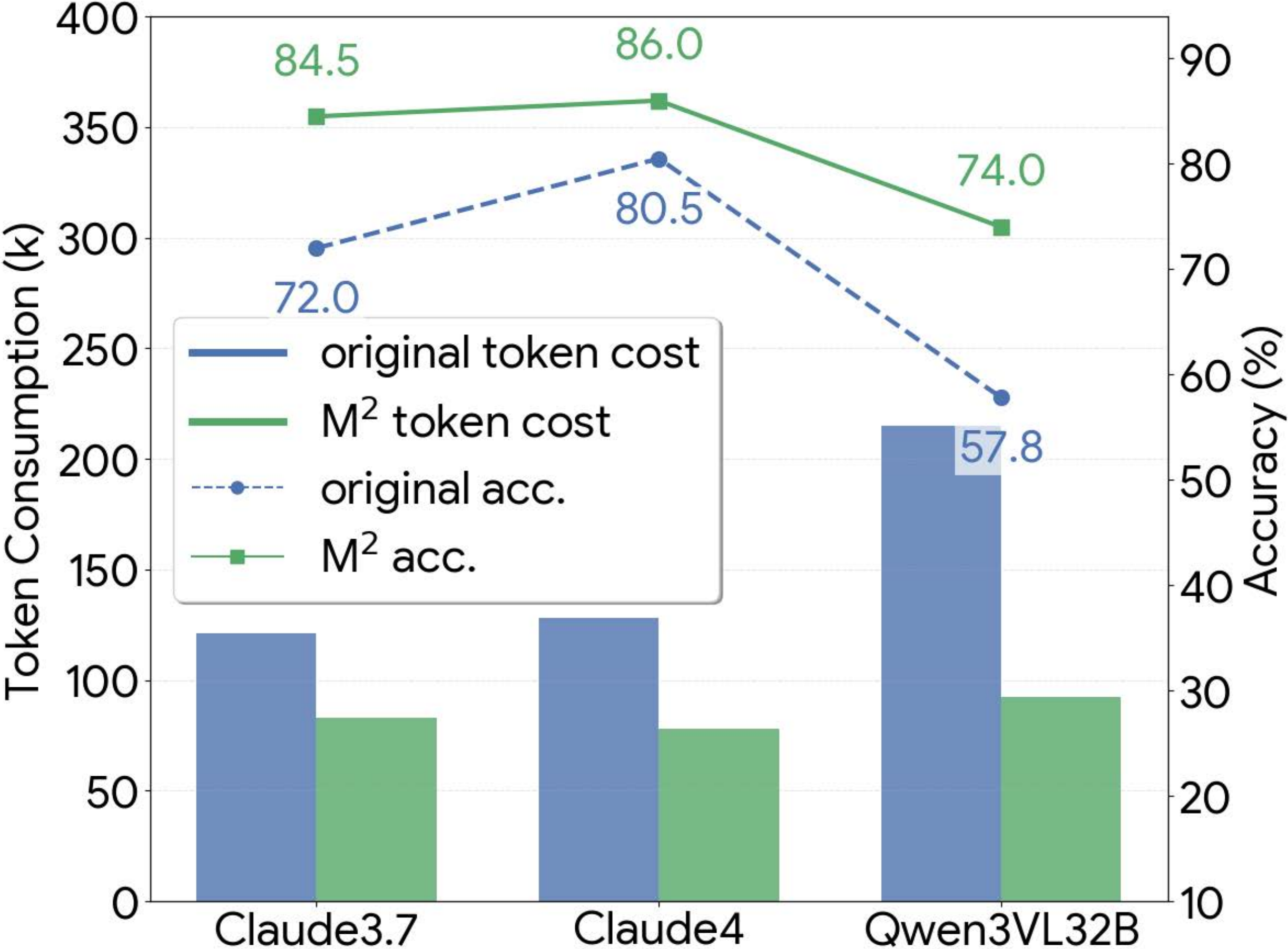}
        \caption{Average Token cost and accuracy of baseline and M$^2$ across models on WebVoyager.}
        \label{fig:visual_example}
    \end{minipage}
\end{figure}

\begin{figure*}[t]
    \centering
    \includegraphics[width=\textwidth]{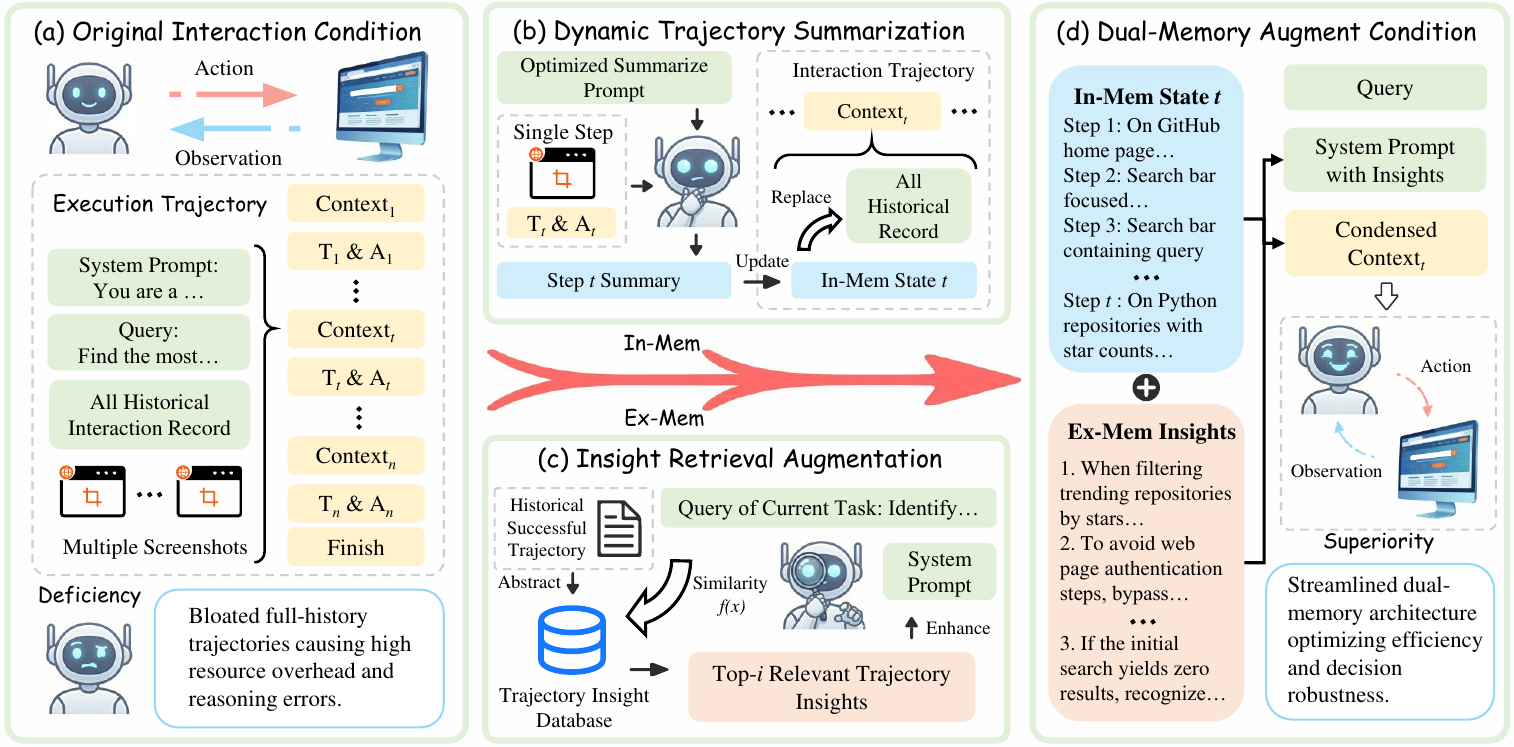}
    \caption{Overview of the proposed framework. (a) The baseline agent operates with raw context containing redundant visual history and verbose interaction text. This creates high computational overhead and introduces noise that may impair decision-making. Our method M$^2$ incorporates two key mechanisms: In-Mem and Ex-Mem. (b) The In-Mem module prompts the agent to self-summarize historical steps into a concise context. Simultaneously, (c) the Ex-Mem module retrieves actionable insights from historically successful trajectories based on query similarity. (d) This dual-memory approach provides explicit guidance for the current execution step, significantly reducing errors and yielding tangible performance gains.}
    \label{fig:method_describe}
\end{figure*}

\section{Introduction}
Recent advancements in MLLMs have catalyzed progress in autonomous web navigation, allowing agents to perform intricate tasks within open-ended web environments~\citep{chen2024webvln,xu2024agenttrek,qi2024webrl,liu2025pal,jain2025augustus,zhang2025litewebagent}. However, real-world web tasks are inherently long-horizon, often requiring numbers of interaction steps to navigate through dynamic web pages, fill forms, and process information. 
Current mainstream approaches typically employ a ``full-context" strategy, concatenating the entire history of HTML screenshots, and interactive text into the prompt~\citep{uitars,li2025websailor,huang2025scaletrack}. This leads to two critical issues: (1) Exorbitant Computational Cost: The accumulation of verbose history and high-resolution screenshots causes the context window to grow rapidly, imposing severe pressure on inference resources. (2) Performance Degradation: An overly long and noisy context often distracts the model, burying critical task-relevant cues under redundant historical information, a phenomenon known as ``lost-in-the-middle"~\citep{liu2024lost}.

% To mitigate the context explosion, recent research has explored memory-augmented mechanisms~\citep{yu2025memagent,wang2025history,cheng2025mga,liu2025webcoach,zhou2025mem1}. One prominent direction involves training the model to autonomously maintain a condensed memory state via Reinforcement Learning (RL) or Supervised Fine-Tuning (SFT). While these methods effectively reduce context length, they introduce considerable drawbacks: (1) High Training Overhead: Constructing trajectories with explicit memory formats and optimizing them via RL requires massive computational resources and delicate reward shaping mechanisms. (2) Loss of Generalization: Agents trained to compress specific web interaction patterns may struggle to generalize to unseen websites or tasks, as the compression policy is tightly coupled with the training distribution. Consequently, a training-free and lightweight approach is highly desirable for scalable web agents. As demonstrated in Table~\ref{tab:sota_sidebyside} , our proposed M$^2$ framework achieves competitive performance compared to state-of-the-art models that rely on extensive SFT and RL, while completely bypassing the associated training overhead.

To mitigate the context explosion, recent research has explored memory-augmented mechanisms~\citep{yu2025memagent,wang2025history,cheng2025mga,liu2025webcoach,zhou2025mem1}. The prominent direction involves training the model to autonomously maintain a condensed memory state via Supervised Fine-Tuning (SFT), Reinforcement Learning (RL), or constructing complex memory architectures composed of multiple collaborative agents. While these methods effectively reduce context length, they introduce considerable drawbacks: (1) Intensive Data and Training Overhead: Optimizing agents through SFT or RL requires massive computational resources, extensive trajectory datasets for training, and delicate reward shaping mechanisms to ensure effective convergence. (2) High Operational and Deployment Costs: Systems relying on multiple agents suffer from substantial overhead in inter-agent communication, increased complexity in deployment, and elevated invocation costs during real-time execution. Consequently, a training-free and lightweight approach is highly desirable for scalable web agents.

Another line of research focuses on external memory banks and graph-based structures to retrieve relevant context. However, directly migrating these general-purpose memory architectures~\citep{wang2025mem,kang2025memory,zhang2025g} to web agents faces distinct domain gaps. First, the interaction nature is different. Unlike conversational agents that need to track multi-turn dialogue history and user personas, web navigation is predominantly execution-oriented. Once the initial instruction is given, there is rarely new user input; tracking detailed interaction history (e.g., every scroll or click) is often redundant unless it contributes to the current subgoal. Second, visual information presents unique challenges. Current multimodal memory systems either store text-only abstractions~\citep{wang2025mirix} or treat visual history as continuous video frames~\citep{he2024ma}. However, web navigation is visually discontinuous—a single click can lead to a completely different page layout, rendering frame-based consistency assumptions invalid. Furthermore, raw screenshots contain excessive visual noise (e.g., ads, sidebars) irrelevant to the task. Storing raw visual history incurs high token costs with low information density, making simple screenshot buffering inefficient.

% To address these challenges, we propose a novel, training-free framework that optimizes context efficiency and execution robustness through a dual-tier memory mechanism. (1) Intra-Trajectory Context Compression (Internal Memory): Instead of relying on expensive training, we utilize a prompt-driven mechanism to dynamically condense the current execution trajectory. By extracting only the essential ``step summaries" and discarding redundant conversation/visual history, we maintain a concise ``working memory" that significantly reduces token consumption without losing track of task progress. (2) Inter-Trajectory Insight Retrieval (External Memory): To enhance success rates, we construct an offline insight bank from historical successful trajectories. Unlike general memory banks, we extract high-level actionable insights (e.g., "use the sidebar filter instead of the search bar for color selection") and retrieve them via semantic similarity to guide the current agent, preventing it from repeating common mistakes. 

% Our approach effectively bridges the gap between cost-efficiency and task performance. By explicitly separating the ``short-term execution state" from ``long-term guidance," we enable the agent to navigate with a lightweight context while benefiting from extensive historical experience.

To address these challenges, we propose a novel, training-free framework that optimizes context efficiency and execution robustness through a dual-tier memory mechanism named M$^2$. As shown in Fig.~\ref{fig:method_describe}, it includes two components: (1) \textbf{Dynamic Trajectory Summarization (Internal Memory)} introduces a prompt-driven self-summarization mechanism. Instead of relying on a sliding window of multiple raw observations that consume excessive tokens, the agent is instructed to distill its own thoughts, actions, and page feedback into concise textual abstractions at each step. By maintaining a recursive chain of these summaries and discarding redundant historical screenshots, we ensure the agent retains a clear sense of progress within a lean context window. Complementing this, (2) \textbf{Insight Retrieval Augmentation (External Memory)} provides global strategic foresight. We construct an offline insight bank by distilling actionable interaction rules and defensive hints from a vast collection of successful trajectories. During inference, the agent retrieves the most relevant insights via semantic similarity based on the user query. This allows the agent to benefit from cross-trajectory experience, effectively navigating complex UI structures and avoiding common pitfalls without any additional training or fine-tuning. As demonstrated in Table~\ref{tab:sota_sidebyside}, our proposed M$^2$ framework achieves competitive performance compared to state-of-the-art models that rely on extensive SFT and RL, while completely bypassing the associated training overhead.

% Comprehensive experiments on WebVoyager and OnlineMind2Web demonstrate that our framework achieves a superior trade-off between execution accuracy and resource efficiency across various model backbones. For the open-source Qwen3-VL-32B, our dual-memory mechanism yields the most significant gains, boosting the success rate by $16.2\%$ on WebVoyager while slashing token consumption by $57.1\%$. On the more challenging OnlineMind2Web benchmark, Qwen achieves an even higher accuracy leap of $19.6\%$ with a $58.8\%$ reduction in tokens. Proprietary models also see substantial benefits: Claude-3.7-Sonnet achieves a $12.5\%$ performance increase on WebVoyager and a $6.9\%$ gain on OnlineMind2Web, while reducing token costs by $30.3\%$ and $55.0\%$ respectively. Similarly, Claude-Sonnet-4 reaches a peak success rate of $86.0\%$ on WebVoyager with a $39.1\%$ token saving, and maintains high efficiency on OnlineMind2Web with a $32.0\%$ cost reduction alongside a $6.2\%$ accuracy improvement. These results confirm that our training-free augmentation not only significantly enhances the navigation capabilities of diverse agents but also ensures sustainable and cost-effective inference for long-horizon web tasks.

Empirical evaluations on WebVoyager~\cite{he2024webvoyager} and OnlineMind2Web~\cite{xue2025illusion} demonstrate that our framework achieves a consistent improvement in both execution success and resource efficiency. The open-source Qwen3-VL-32B~\cite{bai2025qwen3vltechnicalreport} model exhibits the most substantial gains, with success rates surging by 16.2\% on WebVoyager and 19.6\% on OnlineMind2Web, while simultaneously reducing token overhead by over 57\% in both benchmarks. This efficiency trend extends to frontier proprietary models: Claude-3.7-Sonnet~\cite{Claude3S} and Claude-Sonnet-4~\cite{Claude4} achieve accuracy increments of up to 12.5\% alongside significant token reductions ranging from 30.3\% to 55.0\% (see Fig.~\ref{fig:visual_example}). These results validate that our dual-memory architecture effectively decouples task performance from context growth, enabling high-fidelity navigation with sustainable computational costs.

In summary, the major contributions are as follows: 
\begin{itemize}
    \item \textbf{Training-Free Dual-Memory Architecture:} We propose a lightweight framework that integrates recursive internal tracking with external guidance, enabling efficient long-horizon navigation without the need for costly training or cumbersome multi-agent interaction.
    
    \item \textbf{Intra-Trajectory Compression and Inter-Trajectory Retrieval:} We introduce mechanisms to distill execution history into concise summary chains and retrieve cross-task expert insights, effectively mitigating information overload while enhancing decision robustness.
    
    \item \textbf{Scalable Efficacy and Model Parity:} We demonstrate that M$^2$ enables sustainable high-fidelity navigation across diverse and dynamic web domains. 
    % Our approach establishes that training-free mechanisms can bridge the performance gap between open-source and proprietary architectures, empowering local models to achieve parity with leading proprietary counterparts while ensuring operational viability through superior token efficiency.
    Our training-free approach enables local models to match proprietary performance with superior token efficiency.
\end{itemize}

% \begin{itemize} 
%     \item \textbf{Dynamic Trajectory Summarization (Internal Memory):} We introduce a recursive summarization mechanism that dynamically distills verbose execution history into concise state updates. This approach effectively filters out redundant interaction noise while preserving essential task progress, mitigating the ``lost-in-the-middle'' phenomenon.
    
%     \item \textbf{Insight Retrieval Augmentation (External Memory):} We design an external memory module that constructs an offline insight bank from successful historical trajectories. By retrieving and injecting high-level guidelines via semantic similarity, we enable the agent to reference explicit goals and avoid recurring pitfalls during execution.

%     \item \textbf{A Training-Free and Efficient Framework:} We propose a lightweight, model-agnostic architecture that achieves superior performance without the need for expensive fine-tuning or reinforcement learning. Extensive evaluations demonstrate that M$^2$ consistently enhances navigation success rates for both open-source and proprietary models while significantly optimizing token efficiency.
% \end{itemize}

\section{Method: Dual-Memory Augmentation}

\subsection{Overview}
\label{subsec:overview}
In this section, we present the Dual-Memory Augmentation framework M$^2$designed to address the challenges of long-horizon web navigation. 
% Standard Web agents often suffer from context overflow and performance degradation due to the accumulation of historical data. To mitigate this, we reformulate the agent's decision-making process by replacing the raw, exhaustive history with a distilled dual-memory architecture.

\subsubsection{Formalization of Context}
We define a web navigation task as a sequence of steps $t \in \{1, 2, \dots, n\}$, where $n$ denotes the final step of completion. In a conventional setting, at any given step $t$, the agent operates within a context $C_t$. This context typically consists of the system prompt $P_{sys}$, the user query $Q$, and the complete history of previous steps. Each step can be represented as a tuple $S_t = (T_t, A_t)$, $T_t$ is the generated thought (reasoning), and $A_t$ is the executed action. Thus, the vanilla context used in mainstream works is:
% \begin{equation}
%     C_t = \{P_{sys}, Q, S_1, \dots, S_{t-k}, O_{t-k}, \dots, S_{t-1}, O_{t-1} , O_t\}
% \end{equation}
\begin{equation}
\begin{split}
    C_t = \{ & P_{sys}, Q, S_1, S_2, \dots, S_{t-k-1}, \\
             & S_{t-k}, O_{t-k}, \dots, S_{t-1}, O_{t-1} , O_t\}
\end{split}
\end{equation}
where $O_t$ is the page observation with window size $k$.
As $t$ increases, the cardinality of $C_t$ grows indefinitely, leading to high token costs and the ``lost-in-the-middle'' effect. For specific details such as $P_{sys}$, as well as examples of $Q$, $O_t$, and $S_t$, and the action space of the Agent, please refer to Appendix~\ref{sec:prompt_template}.

\subsubsection{The Dual-Memory Framework}
Our approach redefines the context $C_t$ by introducing two specialized memory modules: \textbf{Internal Memory} ($\mathcal{M}^{int}$) and \textbf{External Memory} ($\mathcal{M}^{ext}$). The modified decision-making context at step $t$ is formulated as:
\begin{equation}
    C^{\prime}_t = \{P_{sys}, Q, \mathcal{M}_t^{int}, \mathcal{M}^{ext}, O_t\}
\end{equation}
The interaction between these components is governed by the following logic $(O_{t-k:t-1} = \{O_i\}_{i=t-k}^{t-1})$:

\begin{itemize}
    \item \textbf{Internal Memory $\mathcal{M}_t^{int}$:} This module maintains a dynamic trajectory summary. Instead of retaining the raw observations $O_{t-k:t-1}$ and $S_{1:t-1}$ which are often redundant and noisy, $\mathcal{M}_t^{int}$ stores a chain of concise state abstractions $s_{1:t-1}$. Each $s_t$ is a distilled representation of $(O_t, T_t, A_t)$, ensuring that the internal context remains compact regardless of the task length.
    
    \item \textbf{External Memory $\mathcal{M}^{ext}$:} This module facilitates cross-task experience retrieval. It provides the agent with task-relevant insights $\mathcal{I}$ retrieved from an offline Insight Bank. These insights, denoted as $\mathcal{M}^{ext} = \{ \mathcal{I}_1, \dots, \mathcal{I}_i \}$, serve as global guidance to prevent the agent from repeating historical errors or falling into common pitfalls.
\end{itemize}

By substituting the growing list of historical steps with these two components, our framework ensures that $C_t$ remains within efficient token consumption while providing both local trajectory awareness and global strategic knowledge.

\begin{figure}[t]
    \centering
    \includegraphics[width=0.48\textwidth]{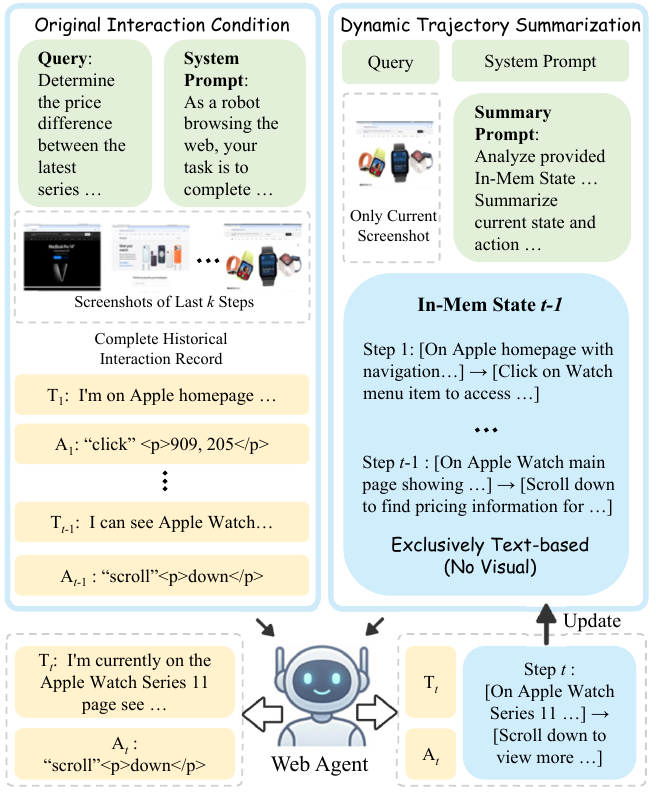}
    \caption{Comparison between Full Context and In Mem interaction paradigms. The Full Context (left) retains the entire sequence of raw screenshots and exhaustive historical text, leading to high redundancy. In contrast, the In Mem approach (right) employs a specific prompt to synthesize historical steps into a concise textual summary.}
    \label{fig:in-mem}
\end{figure}

\subsection{Dynamic Trajectory Summarization}
Typically, HTML pages are replete with task-irrelevant elements—such as advertisements, sidebars, and navigational components unrelated to the current objective. These elements, when captured in full-page screenshots, dominate the visual information. Conversely, interactive targets like search bars or specific buttons often occupy only a negligible fraction of the screen. Consequently, these critical signals are easily overwhelmed by background noise, a persistent issue that is further exacerbated when processing sequences of multiple screenshots.

In contrast, human cognition retains historical interaction information in a much more concise manner. Rather than memorizing the specific visual details of previous pages, humans tend to form a high-level abstraction of their actions (e.g., `I clicked the search bar and entered text'). Inspired by this cognitive mechanism, we propose summarizing historical interaction steps to replace the verbose and potentially distracting visual context, thereby enabling the agent to make more precise decisions.

To maintain a lean context $C_t$ while preserving essential historical information, we propose a Dynamic Trajectory Summarization mechanism. This module replaces the raw, high-dimensional observation sequences with a sequence of semantic abstractions.

\subsubsection{Prompt-Driven State Abstraction}
In our framework, we shift this burden from the raw input buffer to the agent's self-reasoning capability. We augment the system prompt $P_{sys}$ with specific meta-instructions that mandate the agent to perform \textit{self-summarization} alongside its primary decision-making. As shown in Fig.~\ref{fig:in-mem}, the original system prompt is augmented with a new directive that instructs the agent to perform self-summarization. Specifically, at each step $t$, the agent is required to output a triplet $(T_t, A_t, s_t)$ based on the current context:
\begin{equation}
    (T_t, A_t, s_t) = \text{Agent}(P^{in}_{sys}, Q, \mathcal{M}_t^{int}, O_t)
\end{equation}
% where $P^{in}_{sys}$ is the updated system prompt and $\mathcal{M}_t^{int}$ is a concise textual abstraction of the current state transition.
where $P^{in}_{sys}$ is the updated system prompt, and $\mathcal{M}_t^{int}$ is the internal memory consisting of a sequence of concise textual abstractions $s_{1:t-1}$.

By integrating the summarization task directly into the execution loop, the agent treats the abstraction of $s_t$ as a prerequisite for its next action. The instructions within $P^{in}_{sys}$ guide the agent to distill $s_t$ from two perspectives:
\begin{itemize}
    \item \textbf{Visual Feedback:} The observable changes and key information extracted from the page after the action.
    \item \textbf{Action Execution:} A brief record of the action $A_t$ performed on $O_t$.
\end{itemize}

This approach allows the context $C_t$ to retain only the current observation $O_t$. The historical trajectory $\{S_1, S_2, \dots, S_{t-1}\}$ and the multi observations $\{O_{t-k}, \dots, O_{t-1}\}$ is effectively discarded and replaced by the internal memory $\mathcal{M}_t^{int}$ which is constructed by $s_t$. This transformation ensures that the context length grows sub-linearly with the task horizon while maintaining a high density of task-relevant information. For details, please refer to the complete $P^{in}_{sys}$ and $s_t$ samples we provided in Appendix~\ref{sec:prompt_template}.

\subsubsection{Iterative Memory Update}
The core efficiency of our framework lies in the recursive transition of the internal memory state $\mathcal{M}^{int}$, which allows the agent to maintain a persistent state without retaining bulky historical context. Unlike vanilla architectures that store the entire history of $O_{t-k:t-1}$ and $S_{1:t-1}$, our method treats the internal memory as an append-only sequence of distilled summaries.

Formally, at the initialization step $t=1$, the internal memory is empty, i.e., $\mathcal{M}_1^{int} = \emptyset$. For any subsequent step $t > 1$, the internal memory $\mathcal{M}_t^{int}$ is constructed by iteratively appending the summary $s_t$ generated in the previous step to the existing memory state:
\begin{equation}
    \mathcal{M}_t^{int} = [s_1, s_2, \dots, s_{t-1}]
\end{equation}
The update rule for the internal memory state can be expressed as:
\begin{equation}
    \mathcal{M}_t^{int} = \mathcal{M}_{t-1}^{int} \oplus s_{t-1}
\end{equation}
where $\oplus$ denotes the sequence concatenation operator.

% By the time the agent reaches step $t$, the context $C_t$ provided to the LLM is structured as follows:
% \begin{equation}
%     C_t = \{P_{sys}, Q, \mathcal{M}_t^{int}, \mathcal{M}^{ext}, O_t\}
% \end{equation}
Crucially, once $s_{t-1}$ is successfully generated and integrated into $\mathcal{M}_t^{int}$, the observation $O_{t-1}$, the intermediate Thought $T_{t-1}$ and Action $A_{t-1}$ are permanently discarded from the active context window. 

This iterative update mechanism ensures that the memory footprint grows only at the scale of concise text summaries (dozens of tokens) rather than raw webpage observation (typically thousands of tokens). This state-to-state transition effectively transforms long-horizon navigation task into a manageable Markovian-like decision process, where $\mathcal{M}_t^{int}$ serves as a sufficient statistic for historical trajectory.

\begin{figure}[!t]
    \centering
    \includegraphics[width=0.43\textwidth]{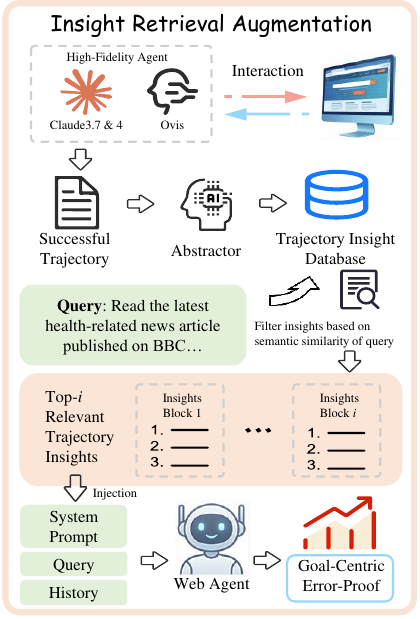}
    \caption{Overview of Insight Retrieval Augmentation pipeline. The workflow begins with a Trajectory Insight Database populated by successful trajectories from diverse models. For a given user query , the system filters insights based on semantic similarity, retrieving the Top-$i$ relevant insights. These insights are then injected  into the System Prompt. This mechanism ensures the Web Agent generates goal-centric and error-proof actions, effectively navigating complex UI environments by leveraging cross-trajectory experiences.}
    \label{fig:ex-mem}
\end{figure}

\subsection{Insight Retrieval Augmentation}
While internal memory ensures trajectory-level coherence, it cannot provide the agent with prior knowledge of common web navigation patterns or potential pitfalls. To bridge this gap, we design an External Memory module that leverages an Insight Bank $\mathcal{B}$ to provide strategic guidance. The complete pipeline is shown in Fig.~\ref{fig:ex-mem}.

% \subsubsection{Offline Insight Extraction from Successful Trajectories}
% The construction of the external memory begins with a heterogeneous collection of successful expert trajectories $\mathcal{T}_{succ}$ generated by various state-of-the-art models, including Claude-3.7, Claude-4, and local Ovis models. To distill high-quality knowledge from these raw logs, we employ the Ovis-2.5-80B model as a dedicated knowledge abstractor. 

% We design a specialized extraction prompt $P_{ext}$ that instructs the Ovis model to analyze the transition from the initial query $Q$ to the final success state. The extractor focuses on identifying:
% \begin{itemize}
%     \item \textbf{Operational Heuristics:} Key sequences of actions that efficiently lead to the goal.
%     \item \textbf{Pitfall Avoidance:} Counter-intuitive behaviors or common errors (e.g., incorrect coordinate mappings, misleading pop-ups) that were successfully bypassed in the trajectory.
% \end{itemize}
% For each trajectory $i$, the extraction result is a concise set of atomic insights $\mathcal{I}$.

\subsubsection{Offline Insight Extraction from Successful Trajectories}
The efficacy of our external memory depends on the quality of the distilled insights. To transform raw interaction logs into transferable strategic knowledge, we implement a systematic reverse-engineering process. We utilize a heterogeneous collection of successful trajectories $\mathcal{T}_{succ}$ generated by various frontier models (e.g., Claude-3.7, Claude-4, and Ovis~\cite{lu2024ovisstructuralembeddingalignment,lu2025ovis25technicalreport}) as the source data.

To perform high-fidelity knowledge distillation, we employ the Ovis-inhouse 80A3B model as a Web Navigation Strategy Abstractor. This model is guided by a specialized extraction prompt $P_{ext}$ designed to identify ``High-Leverage Interaction Rules'' from successful trajectories. It is worth noting that our framework is model-agnostic; Ovis is selected here primarily for its favorable balance between performance and deployment efficiency, serving as a representative instantiation of a capable vision-language model. The extraction process is formulated as:
\begin{equation}
    \mathcal{B} = \text{Abstractor}(\mathcal{T}_{succ}, P_{ext})
\end{equation}
The prompt $P_{ext}$ enforces a focus on pure UI interaction logic, strictly prohibiting URL manipulation and trivial action reporting. It requires the abstractor to distill insights across three critical dimensions:

\begin{itemize}
    \item \textbf{Search \& Filter Strategy (Input Logic):} This dimension focuses on the construction of precision queries. It identifies optimal keyword prefixes, the use of exact match operators, and the strategic combination of UI filters to narrow down the search space immediately.
    \item \textbf{Navigation Efficiency (Click Logic):} The extractor identifies ``Anchor Elements''—specific UI components (e.g., advanced search buttons or hierarchical breadcrumbs) that serve as shortcuts to avoid deep, irrelevant navigation paths and prevent the agent from wandering.
    \item \textbf{State Validation (Wait/Check Logic):} This provides agent with robustness heuristics, such as verifying UI state changes or implementing fallback strategies when a specific query yields no results, thereby avoiding infinite loops and traps.
\end{itemize}

Crucially, the extraction prompt requires the generalization of task-specific values into universal strategies. For example, specific item names are abstracted into ``target items'' to ensure the resulting insights $\mathcal{I}$ are applicable to a broad range of semantically similar tasks. The output is formatted as a set of topic-tagged, actionable insights, providing a structured knowledge base for the subsequent retrieval phase. We provide the complete $P_{ext}$ in Appendix~\ref{sec:prompt_template}, as well as the distribution of these trajectories and an insight example in Appendix~\ref{sec:insight_bank}.

\subsubsection{Construction of Insight Bank}
The insights extracted are organized into a key-value store, referred to as the Insight Bank $\mathcal{B}$. Each entry in the bank is represented as a pair $(Q^{hist}, \mathcal{I})$, where $Q^{hist}$ is the historical user query that initiated the successful trajectory, and $\mathcal{I}$ is the corresponding insights. 

To enable semantic retrieval, we utilize a pre-trained Sentence Transformer model $\phi(\cdot)$ to encode the historical queries into a high-dimensional vector space:
\begin{equation}
    \mathbf{v}^{hist} = \phi(Q^{hist})
\end{equation}
The set of $N$ trajectories in bank can be characterized as $\{(\mathbf{v}_j^{hist}, \mathcal{I})\}_{j=1}^N$ thus serves as a global experience repository that the agent can query during inference.

\subsubsection{Similarity-Based Retrieval and Contextual Injection}
During execution of a new task with query $Q^{new}$, agent first computes the query's embedding $\mathbf{v}^{new} = \phi(Q^{new})$. We then calculate the semantic similarity between current query and all entries in $\mathcal{B}$ using cosine similarity:
\begin{equation}
    \text{sim}(Q^{new}, Q^{hist}) = \frac{\mathbf{v}^{new} \cdot \mathbf{v}_j^{hist}}{\|\mathbf{v}^{new}\| \|\mathbf{v}_j^{hist}\|}
\end{equation}
The $i$ most similar queries are identified, and their corresponding insights are retrieved to form the external memory $\mathcal{M}^{ext} = \{\mathcal{I}_1, \dots, \mathcal{I}_i\}$.

Finally, these insights are injected into the system prompt $P_{sys}$ as ``Defensive Hints.'' The agent is instructed to treat these insights as strategic suggestions that must be verified against the current observation $O_t$ before execution. This injection transforms the agent from a purely reactive entity into a goal-oriented navigator that actively avoids past mistakes, thereby significantly increasing the success rate in complex, long-horizon tasks.

\begin{table*}[t]
\centering
\caption{Performance Comparison of Claude and Qwen3VL Model across Multiple WebSite on WebVoyager. (Abbreviations: AR: Allrecipes, AM: Amazon, AP: Apple, AX: ArXiv, BN: BBC News, CD: Cambridge Dict., CO: Coursera, ES: ESPN, GH: GitHub, GM: Google Map, HF: Huggingface, WA: Wolfram Alpha.)}
\label{tab:model_comparison}
\resizebox{\textwidth}{!}{ 
\begin{tabular}{lccccccccccccc|ccc}
\toprule
\textbf{Model} & \textbf{Config} & \textbf{AR} & \textbf{AM} & \textbf{AP} & \textbf{AX} & \textbf{BN} & \textbf{CD} & \textbf{CO} & \textbf{ES} & \textbf{GH} & \textbf{GM} & \textbf{HF} & \textbf{WA} & \textbf{Avg.step} & \textbf{Avg.token} & \textbf{Avg.acc.} \\ 
\midrule
\multirow{4}{*}{Claude-3.7-Sonnet} 
 & Normal & 71.1 & 65.9 & 69.8 & 74.4 & 78.6 & 83.7 & 83.3 & 77.3 & 75.6 & 63.4 & 44.2 & 76.1 & 14.6 & 121.2k & 72.0 \\
 & In Mem & 73.3 & 70.7 & 69.8 & 62.8 & 78.6 & 88.4 & 83.3 & 68.2 & 78.1 & 63.4 & 58.1 & 80.4 & 20.3 & 82.9k & 72.9 \\
 & Ex Mem & 82.2 & 78.1 & 76.7 & 76.7 & 83.3 & 86.1 & 85.7 & 68.2 & 90.2 & 80.5 & 67.4 & 82.6 & 15.7 & 164.9k & 79.8 \\
 & In \& Ex Mem & 84.4 & 80.5 & 76.7 & 86.1 & 85.7 & 93.0 & 95.2 & 84.1 & 92.7 & 82.9 & 74.4 & 78.3 & 16.9 & 84.5k & \textbf{84.5} \gain{$+$12.5\%} \\ 
\midrule
\multirow{4}{*}{Claude-Sonnet-4} 
 & Normal & 82.2 & 70.7 & 69.8 & 79.1 & 83.3 & 93.0 & 85.7 & 84.1 & 87.8 & 78.1 & 69.8 & 82.6 & 15.4 & 128.2k & 80.5 \\
 & In Mem & 84.4 & 75.6 & 79.1 & 81.4 & 78.6 & 88.4 & 85.7 & 79.6 & 82.9 & 70.7 & 67.4 & 82.6 & 16.8 & 64.5k & 79.7 \\
 & Ex Mem & 84.4 & 85.4 & 83.7 & 83.7 & 92.9 & 86.1 & 70.5 & 85.4 & 85.4 & 73.2 & 72.1 & 89.1 & 12.4 & 120.7k & 82.7 \\
 & In \& Ex Mem & 82.2 & 78.1 & 81.4 & 81.4 & 85.7 & 95.4 & 92.9 & 95.5 & 92.7 & 87.8 & 74.4 & 84.8 & 16.1 & 78.1k & \textbf{86.0} \gain{$+$5.5\%} \\ 
\midrule
\multirow{4}{*}{Qwen3-VL-32B} 
 & Normal & 55.6 & 51.2 & 60.5 & 46.5 & 73.8 & 83.7 & 71.4 & 47.7 & 56.1 & 56.1 & 30.2 & 60.9 & 24.1 & 215.2k & 57.8 \\
 & In Mem & 62.2 & 75.6 & 60.5 & 44.2 & 73.8 & 76.7 & 76.2 & 70.5 & 80.5 & 58.5 & 37.2 & 63.0 & 22.5 & 73.2k & 64.9 \\
 & Ex Mem & 71.1 & 63.4 & 67.4 & 65.1 & 83.3 & 88.4 & 81.0 & 63.6 & 70.7 & 73.2 & 46.5 & 67.4 & 19.2 & 189.1k & 70.1 \\
 & In \& Ex Mem & 75.6 & 70.7 & 72.1 & 72.1 & 90.5 & 88.4 & 81.0 & 77.3 & 80.5 & 68.3 & 48.8 & 63.0 & 21.8 & 92.3k & \textbf{74.0} \gain{$+$16.2\%} \\ 
\bottomrule
\end{tabular}
}
\end{table*}

\subsection{Dual-Memory Integration and Inference}
The final decision-making architecture of our agent is a synthesis of local trajectory awareness and global strategic knowledge. By integrating the internal and external memory modules, we reconstruct the operational context $C_t$ at each time step $t$. This integration allows the agent to navigate complex web environments with both hindsight (from past steps) and foresight (from expert insights).

\subsubsection{Synthesized Decision Context}
As established in the preceding sections, the vanilla context—which suffers from rapid growth in length—is replaced by our dual-memory augmented context $C'_t$. For any step $t \in \{1, \dots, n\}$, the agent generates the current thought $T_t$, action $A_t$, and state summary $s_t$ based on the following synthesized formulation:
\begin{equation}
    (T_t, A_t, s_t) = \text{Agent}(P^{\prime}_{sys}, Q, \mathcal{M}_t^{int}, \mathcal{M}^{ext}, O_t)
\end{equation}
where:
\begin{itemize}
    \item $\mathcal{M}_t^{int} = \{s_1, \dots, s_{t-1}\}$ provides \textbf{local trajectory awareness}, ensuring agent maintains a compact yet coherent record of its progress within current session.
    \item $\mathcal{M}^{ext} = \text{Top$_i$}(\phi(Q), \mathcal{B})$ provides \textbf{global heuristic guidance}, injecting task-specific ``defensive hints'' retrieved from the offline Insight Bank to prevent the recurrence of historical failures.

    \item $P^{\prime}_{sys}$ represents the optimized $P_{sys}$ that fits M$^2$ framework. Please refer to Appendix~\ref{sec:prompt_template} for complete $P^{\prime}_{sys}$.
\end{itemize}

\subsubsection{Synergy and Efficiency}
The interplay between $\mathcal{M}_t^{int}$ and $\mathcal{M}^{ext}$ creates a powerful synergy for long-horizon web navigation. While $\mathcal{M}_t^{int}$ anchors agent to current task's state transitions, $\mathcal{M}^{ext}$ acts as a regularizer, correcting agent's trajectory when it encounters ambiguous page elements or complex interaction logic.

From an efficiency perspective, this dual-memory structure effectively decodes the long-horizon dependency problem into a sub-linearly growing context window. By discarding raw historical context $O_{t-k:t-1}$ \& $S_{1:t-1}$ and retaining only the distilled summaries and Top-$i$ insights, our framework ensures that the token consumption remains stable even as the task depth increases. Consequently, this architecture achieves a superior trade-off between computational cost and task success rate, fulfilling the primary objective of our training-free augmentation approach.

\begin{table*}[t]
\centering
\caption{Evaluation Results on OnlineMind2Web. Green values in parentheses represent the relative reduction percentage for Tokens and absolute percentage point increase for Accuracy.}
\label{tab:final_version}
\resizebox{\textwidth}{!}{
\begin{tabular}{ll ccc ccc ccc ccc}
\toprule
\multirow{2}{*}{\textbf{Model}} & \multirow{2}{*}{\textbf{Config}} & \multicolumn{3}{c}{\textbf{Easy}} & \multicolumn{3}{c}{\textbf{Medium}} & \multicolumn{3}{c}{\textbf{Hard}} & \multicolumn{3}{c}{\textbf{Average}} \\
\cmidrule(lr){3-5} \cmidrule(lr){6-8} \cmidrule(lr){9-11} \cmidrule(lr){12-14}
 & & Step & Token & Acc. (\%) & Step & Token & Acc. (\%) & Step & Token & Acc (\%) & Step & Token & Acc. (\%) \\
\midrule
\multirow{2}{*}{Claude-3.7-Sonnet} 
 & Normal & 21.72 & 219.6k & 74.07 & 29.50 & 310.9k & 58.27 & 34.51 & 371.6k & 53.52 & 28.56 & 300.3k & 61.95 \\
 & In \& Ex & 19.89 & 98.0k \gain{$-$55.4\%} & 79.01 \gain{+4.9\%} & 27.40 & 144.0k \gain{$-$53.7\%} & 65.47 \gain{+7.2\%} & 31.03 & 163.9k \gain{$-$55.9\%} & 61.97 \gain{+8.5\%} & 26.05 & 135.2k \gain{$-$55.0\%} & 68.82 \gain{+6.9\%} \\
\midrule
\multirow{2}{*}{Claude-Sonnet-4} 
 & Normal & 16.23 & 154.7k & 74.07 & 21.24 & 205.9k & 66.91 & 21.24 & 292.1k & 61.97 & 21.54 & 212.7k & 67.70 \\
 & In \& Ex & 18.77 & 95.7k \gain{$-$38.1\%} & 82.72 \gain{+8.7\%} & 27.00 & 141.2k \gain{$-$31.4\%} & 73.38 \gain{+6.5\%} & 37.24 & 207.2k \gain{$-$29.1\%} & 64.79 \gain{+2.8\%} & 27.21 & 144.6k \gain{$-$32.0\%} & 73.88 \gain{+6.2\%} \\
\midrule
\multirow{2}{*}{Qwen3-VL-32B} 
 & Normal & 25.41 & 220.9k & 44.44 & 35.38 & 325.8k & 30.22 & 42.11 & 403.7k & 21.13 & 34.25 & 315.6k & 31.96 \\
 & In \& Ex & 24.67 & 104.0k \gain{$-$52.9\%} & 65.43 \gain{+21.0\%} & 29.67 & 127.1k \gain{$-$61.0\%} & 52.52 \gain{+22.3\%} & 37.21 & 166.0k \gain{$-$58.9\%} & 33.80 \gain{+12.7\%} & 30.12 & 130.2k \gain{$-$58.7\%} & 51.55 \gain{+19.6\%} \\
\bottomrule
\end{tabular}
}
\end{table*}

\section{Experiment}
In this section, we evaluate the performance of our dual-memory augmented framework across diverse real-time web navigation benchmarks. We describe the experimental configurations, evaluation protocols, and the infrastructure used to simulate real-world user interactions.

\subsection{Basic Settings}

\textbf{Datasets and Models.}
Our framework is evaluated on two benchmarks: WebVoyager (12 representative websites) and OnlineMind2Web (291 accessible tasks). We employ a diverse set of multimodal backbones as agents, including Claude 3.7 \& 4 computer-use version, and open-source Qwen3-VL-32B. 

\textbf{Infrastructure and Constraints.}
All experiments are executed on a high-concurrency Kubernetes cluster. Agents interact with virtualized browser environments via kubectl. Each task is restricted to a maximum of 60 interaction steps. While the baseline utilizes a raw observation window of size $k=5$, our dual-memory approach optimizes this into a sub-linear summary chain.

\textbf{Implementation and Evaluation.}
% The external Insight Bank is constructed from 55k successful trajectories, involving websites distributed along with WebVoyager. 
The external Insight Bank is constructed from 55k successful trajectories; while they share the same website distribution as WebVoyager, they involve a distinct set of user queries.
For each query, we retrieve the Top-5 insights using sentence transformer and cosine similarity. Following the interaction, GPT-4o~\cite{hurst2024gpt} determines task success by analyzing the agent's textual response and last five screenshots of the trajectory. More details on configurations are provided in Appendix~\ref{sec:implementation}.

\subsection{Main Results and Analysis}

\subsubsection{Overall Performance and Generalization}
The experimental results on WebVoyager and OnlineMind2Web, summarized in Table~\ref{tab:model_comparison} and Table~\ref{tab:final_version}, demonstrate the consistent superiority of our Dual-Memory Augmentation framework across various model backbones and task complexities. We provide several examples of visualized agent interaction trajectories with corresponding analysis in Appendix~\ref{sec:qual_analysis}.

\textbf{Overall Superiority.} In all tested scenarios, the combined ``In \& Ex Mem'' configuration achieves the highest success rates. On WebVoyager, Claude-3.7-Sonnet and Claude-Sonnet-4 reach success rates of $84.5\%$ and $86.0\%$ respectively, while the open-source Qwen3-VL-32B sees a significant jump from $57.8\%$ to $74.0\%$. Notably, our framework effectively bridges the gap between open-source and closed-source models; with dual-memory augmentation, the local Qwen3-VL-32B model ($74.0\%$) even outperforms the vanilla Claude-3.7-Sonnet agent ($72.0\%$), highlighting that robust memory management can compensate for limitations in web navigation.

\textbf{Robustness Across Task Complexity.} Analysis of the OnlineMind2Web dataset (Table~\ref{tab:final_version}) reveals that the performance gains are most pronounced in more challenging scenarios. In ``Medium'' tasks, the success rate of Qwen3-VL-32B increases from $30.22\%$ to $52.52\%$, a relative improvement of nearly $73\%$. This suggests that our memory mechanism provides crucial stability when the agents are faced with the deep navigation hierarchies and the complex UI interaction chains.

\textbf{Cross-Trajectory Knowledge Transfer.} A key finding is the efficacy of our Insight Bank. 
% Although the insights are primarily extracted from successful trajectories with a distribution of website similar to WebVoyager, they demonstrate a universal value across different benchmarks and models. 
Although the trajectories align with the website distribution of WebVoyager, the extracted insights demonstrate a universal value across different benchmarks and models.
Even when applied to the dynamic OnlineMind2Web environment, these insights provide strategic guidance that translates into immediate success rate improvements for Claude and Qwen3VL models alike, proving that high-level website UI interaction rules are transferable across different web domains.

\subsubsection{Efficiency and Practicality Analysis}
Beyond raw success rates, we investigate the underlying mechanisms of efficiency and decision-making quality provided by our internal and external memory modules.

\textbf{Efficiency Gains via Internal Memory.} As shown in the ``Avg.token'' columns, the introduction of internal memory ($\mathcal{M}^{int}$) leads to a dramatic reduction in computational cost. By replacing redundant raw observation windows with concise trajectory summaries, the average token consumption per task for Qwen3-VL-32B on WebVoyager drops from $215.2k$ to $92.3k$, a saving of approximately $57\%$. This confirms that our training-free framework effectively mitigates the context-window pressure inherent in long-horizon tasks without sacrificing historical context.

\textbf{Practicality via External Memory.} The ablation of memory modules (Table~\ref{tab:model_comparison}) illustrates that external memory ($\mathcal{M}^{ext}$) is the primary driver of decision practicality. For instance, in the Google Maps domain, the accuracy of Claude-3.7-Sonnet increases from $63.4\%$ to $80.5\%$ with the addition of insights. These ``defensive hints'' prevent the agent from repeating common pitfalls and ensure that navigation remains goal-centric even in visually noisy environments. Furthermore, performance profiling indicates that against an average task duration of 122 seconds on WebVoyager, the latency for a single similarity-based insight retrieval is approximately 6 milliseconds, demonstrating that the computational overhead introduced by $\mathcal{M}^{ext}$ is negligible and highly sustainable for real-time deployment.

\textbf{Trade-off between Steps and Accuracy.} We observe an interesting correlation between average steps and accuracy. In certain configurations, the average steps slightly increase alongside accuracy (e.g., Claude-3.7-Sonnet on WebVoyager). This is not an indication of inefficiency, but rather a reflection of the agent's enhanced persistence. M$^2$ provides the agent with a clearer sense of progress and strategic guidance, preventing the model from prematurely judging the task as uncompletable and exiting. By maintaining a more stable and compact context, the agent is empowered to explore longer, correct its own course, and eventually reach the target state. We visualize the token consumption of the baseline and M$^2$ on a specific task in Appendix~\ref{sec:token_analysis}.

\begin{table}[htbp]
\centering
% \footnotesize % 减小字号
% \setlength{\tabcolsep}{3pt} % 压缩列间距
% \renewcommand{\arraystretch}{0.9} % 略微减小行间距
\caption{Ablation study of M$^2$ components on WebVoyager by Qwen3-VL-32B. S.T. denotes Sentence Transformer (Default).}
\label{tab:ablation_study}
\resizebox{0.48\textwidth}{!}{
\begin{tabular}{lcc}
\toprule
\textbf{Ablation Category} & \textbf{Settings} & \textbf{Acc. (\%)} \\
\midrule
\multirow{4}{*}{Historical Window $k$} & 1 & 51.6  \\
 & 3 & 56.0  \\
 & 5 (Default) & \textbf{57.8}  \\
  & 7 & 52.2  \\
\midrule
\multirow{4}{*}{Similarity Function} & TF-IDF~\cite{salton1988term} & 61.1  \\
 & Word2Vec~\cite{mikolov2013efficient} & 64.4  \\
 & BM25~\cite{robertson1995okapi} & 58.0  \\
 & S.T.~\cite{reimers-2019-sentence-bert} & \textbf{70.1}  \\
\midrule
\multirow{3}{*}{Insight Count $i$} & 3 & 64.8  \\
 & 5 (Default) & \textbf{70.1}  \\
 & 7 & 66.2  \\
\midrule
\multirow{3}{*}{In-Mem Components} & w/o Visual & 62.3  \\
 & w/o Action & 54.7  \\
 & Full & \textbf{64.9}  \\
\bottomrule
\end{tabular}}
% \vspace{-0.5cm}
\end{table}

\subsection{Ablation Studies}
% We perform ablation experiments on the specific settings of M$^2$ framework to evaluate parameter influence on agent performance. Expanding the historical window to five steps and utilizing sentence transformers significantly improves success by providing essential temporal context and capturing semantic intent. Optimal guidance density is achieved with five insights, whereas higher counts distract the agent with redundant information. Finally, while visual feedback remains critical, removing action records causes the most significant accuracy decline, identifying them as the primary element for maintaining trajectory continuity.
We perform ablation experiments on the specific settings of the M$^2$ framework to systematically evaluate the influence of various parameters on agent performance. (1) Regarding the historical observation window size, performance scales with context depth up to five steps, demonstrating that while moderate temporal cues are essential for resolving state ambiguity, excessive historical length introduces contextual interference that can degrade agent reasoning. (2) In the evaluation of similarity functions, the sentence transformer outperforms traditional lexical methods such as TF-IDF and BM25, indicating that capturing the underlying semantic intent of a query is more effective than simple keyword overlap. (3) For insight retrieval count, five insights provide optimal density of guidance; increasing the number to seven leads to a performance decline, likely due to the introduction of redundant information that distracts the agent. (4) Finally, the analysis of internal memory components demonstrates that while both visual feedback and action history are critical, the removal of action records causes a more significant drop in accuracy, identifying it as the primary element for maintaining trajectory continuity and ensuring robust decision-making. 
% Collectively, these findings validate the architectural hypothesis of M$^2$: that agent performance is maximized not by simply extending context length, but by optimizing the \textit{quality} and \textit{relevance} of context. 
% This ``precision-over-volume'' approach ensures that the agent remains focused on high-value cues without succumbing to the cognitive overload often observed in full-context baselines.

\section{Conclusion}
% In this paper, we introduce the Dual-Memory framework M$^2$, a training-free approach designed to enhance the execution robustness and computational efficiency of web agents. By integrating Intra-Trajectory Context Compression with Inter-Trajectory Insight Retrieval, we address the dual challenges of context explosion and performance degradation in long-horizon tasks. Extensive evaluations on WebVoyager and OnlineMind2Web demonstrate that our framework significantly improves navigation success across diverse model backbones. Most notably, the open-source Qwen3-VL-32B achieved a 16.2\% increase in success rate alongside a 57\% reduction in token consumption, effectively bridging the performance gap with proprietary models. These results establish that the dual-memory architecture provides a scalable and sustainable path for deploying high-fidelity web agents in real-world environments.

In this paper, we introduce the Dual-Memory framework M$^2$, a training-free approach designed to enhance the execution robustness and computational efficiency of autonomous web agents. 
% By integrating Dynamic Trajectory Summarization and Insight Retrieval Augmentation, we effectively address the dual challenges of context explosion and performance degradation in long-horizon tasks. 
Our recursive summarization mechanism distills interaction history into concise state updates, while the retrieval of offline expert insights provides the strategic guidance necessary to navigate complex user interfaces without the overhead of data production or training.
Extensive evaluations on WebVoyager and OnlineMind2Web demonstrate that our framework significantly improves navigation success across diverse model backbones. Most notably, the open-source Qwen3-VL-32B achieved a 16.2\% increase in success rate alongside a 57\% reduction in token consumption, effectively bridging the performance gap with proprietary models. These results establish that the dual-memory architecture provides a scalable and sustainable path for deploying high-fidelity web agents in real-world environments.

\section*{Impact Statement}

This paper presents work whose goal is to advance the field of Machine Learning, specifically in the domain of autonomous web navigation agents. There are many potential societal consequences of our work, none which we feel must be specifically highlighted here.
% \section*{Impact Statement}

% Authors are \textbf{required} to include a statement of the potential broader
% impact of their work, including its ethical aspects and future societal
% consequences. This statement should be in an unnumbered section at the end of
% the paper (co-located with Acknowledgements -- the two may appear in either
% order, but both must be before References), and does not count toward the paper
% page limit. In many cases, where the ethical impacts and expected societal
% implications are those that are well established when advancing the field of
% Machine Learning, substantial discussion is not required, and a simple
% statement such as the following will suffice:

% ``This paper presents work whose goal is to advance the field of Machine
% Learning. There are many potential societal consequences of our work, none
% which we feel must be specifically highlighted here.''

% The above statement can be used verbatim in such cases, but we encourage
% authors to think about whether there is content which does warrant further
% discussion, as this statement will be apparent if the paper is later flagged
% for ethics review.


\bibliography{example_paper}
\bibliographystyle{icml2026}

%%%%%%%%%%%%%%%%%%%%%%%%%%%%%%%%%%%%%%%%%%%%%%%%%%%%%%%%%%%%%%%%%%%%%%%%%%%%%%%
%%%%%%%%%%%%%%%%%%%%%%%%%%%%%%%%%%%%%%%%%%%%%%%%%%%%%%%%%%%%%%%%%%%%%%%%%%%%%%%
% APPENDIX
%%%%%%%%%%%%%%%%%%%%%%%%%%%%%%%%%%%%%%%%%%%%%%%%%%%%%%%%%%%%%%%%%%%%%%%%%%%%%%%
%%%%%%%%%%%%%%%%%%%%%%%%%%%%%%%%%%%%%%%%%%%%%%%%%%%%%%%%%%%%%%%%%%%%%%%%%%%%%%%

\clearpage
\appendix

% \onecolumn

\section{Related Work}
\subsection{Context Management and Memory Optimization in LLM Agents}
Effective context management is a fundamental prerequisite for deploying LLM based agents in long-horizon web navigation tasks. 
% As agents interact with dynamic environments, the accumulation of historical observations—including verbose HTML structures, accessibility trees, and sequential actions—results in an exponentially growing context. 
% This not only incurs prohibitive computational costs but also exacerbates the 'lost-in-the-middle' phenomenon, where critical task-relevant information is overshadowed by noise. 
Recent literature has extensively explored strategies to optimize memory representation and context utilization~\citep{ge2025tremu,tan2025membench,pink2025position,xu2025mem,zhang2025memengine}. 
% MemGPT~\citep{packer2023memgpt} proposes a virtual context management technique that intelligently manages data movement between fast (context window) and slow (external storage) memory tiers, effectively enabling LLMs to handle unbounded contexts.
% MIRIX~\citep{wang2025mirix} introduces a modular, multi-agent framework that orchestrates six distinct memory types (including episodic, semantic, and procedural) to enable persistent reasoning and scalable context management over long-term user interactions. VideoAgent~\citep{fan2024videoagent} constructs a unified structured memory that retains both generic temporal event descriptions and object-centric tracking states, enabling LLMs to effectively manage and query extensive visual contexts. To emulate human-like memory accumulation, M3-Agent~\citep{long2025seeing} introduces an entity-centric multimodal memory framework that dynamically updates episodic and semantic memories from real-time visual and auditory inputs, facilitating deeper environmental consistency. 
% MGA~\citep{cheng2025mga} proposes a memory-driven framework that models GUI interactions as independent states—supported by a dynamically updated structured memory—to decouple decision-making from long-chain dependencies.
To manage extensive contexts, MemGPT~\citep{packer2023memgpt} utilizes a tiered storage mechanism to handle unbounded data, while MIRIX~\citep{wang2025mirix} orchestrates six distinct memory types for persistent reasoning. In multimodal domains, VideoAgent~\citep{fan2024videoagent} and M3-Agent~\citep{long2025seeing} employ structured, entity-centric frameworks to track visual and auditory states dynamically. Similarly, MGA~\citep{cheng2025mga} leverages structured memory to model GUI interactions as independent states, effectively decoupling long-chain dependencies.

% Despite the advancements in extending context capacity and compressing information, existing methods face significant limitations when applied to web agents. First, token-level or embedding-based compression often disrupts the structural integrity of web elements (e.g., DOM IDs), rendering the compressed context unexecutable for agents. Second, while learning-based summarization (e.g., via RL) preserves semantic coherence, it demands substantial training overhead and often overfits to specific task distributions. Addressing these limitations, our work proposes a training-free, structure-aware trajectory summarization mechanism. Instead of implicit compression or expensive training, we leverage prompt-driven abstraction to dynamically maintain a concise state that filters visual noise while strictly preserving actionable navigation cues.
Current context strategies typically suffer from two drawbacks: the reliance on decoupled summarization agents which increases inference cost, and the dependence on learning-based models that require extensive training. We overcome these by proposing a lightweight, training-free alternative. Our method employs in-context prompt abstraction to filter visual noise and retain navigation cues, eliminating the need for heavy external modules or parameter updates.

\subsection{Experience Learning and Retrieval-Augmented Generation for LLM Agents}
By leveraging paradigms like Experience Learning or Retrieval-Augmented Generation (RAG) to retrieve relevant insights, agents can transcend immediate context limitations to avoid recurring pitfalls and enhance generalization~\cite{salama2025meminsight,wheeler2025procedural,rasmussen2025zep}.
% ICAL~\citep{sarch2024vlm} proposes a framework that transforms suboptimal multimodal trajectories into high-quality cognitive abstractions—such as causal relationships and temporal subgoals—via self-reflection and human feedback, significantly enhancing performance in retrieval-augmented setups. Optimus-1~\citep{li2024optimus} utilizes an Abstracted Multimodal Experience Pool that summarizes historical interactions into rich references for in-context learning, coupled with an experience-driven reflector to enhance planning capabilities. RMM~\citep{tan2025prospect} incorporates a retrospective reflection module that iteratively refines memory retrieval using online reinforcement learning based on the evidence cited by the model, thereby optimizing performance in open-ended dialogues. G-Memory~\citep{zhang2025g} implements a bi-directional retrieval mechanism that assimilates cross-trial knowledge by combining high-level generalizable insights with fine-grained interaction trajectories, significantly improving success rates in embodied tasks and QA. HAR~\cite{wang2025history} proposes a history-aware framework that employs a reflective learning scenario and hybrid reinforcement learning rewards to encourage agents to reflect on past errors and synthesize tailored correction guidelines for long-horizon tasks.
Focusing on abstraction and reflection, ICAL~\citep{sarch2024vlm} transforms suboptimal trajectories into cognitive abstractions via self-correction, while Optimus-1~\citep{li2024optimus} utilizes an experience pool of summarized interactions to enhance planning. Incorporating reinforcement learning, RMM~\citep{tan2025prospect} iteratively refines memory retrieval using online RL, and HAR~\cite{wang2025history} employs hybrid rewards to synthesize error correction guidelines. Furthermore, G-Memory~\citep{zhang2025g} integrates high-level insights with fine-grained trajectories through a bi-directional retrieval mechanism.

% However, most existing experience learning frameworks focus predominantly on textual reasoning or open-domain question answering, lacking specificity for the complex dynamics of web navigation. On one hand, general RAG retrieval often fetches factual knowledge rather than procedural actionable insights (e.g., how to interact with a specific filter), which are crucial for web agents. On the other hand, reflection-based methods often require online interaction or model fine-tuning to internalize experiences, which is computationally expensive and difficult to scale. To bridge this gap, we introduce a semantic Insight Retrieval Augmentation module. By constructing an offline bank of high-level insights extracted from successful trajectories and retrieving them via task similarity, our approach provides the agent with robust, interpretable guidance without the need for additional parameter updates.
Existing frameworks primarily target textual tasks, often lacking the procedural specificity required for web navigation. Moreover, they typically rely on factual retrieval or online interactions. Addressing these limitations, we propose a semantic Insight Retrieval Augmentation module. This mechanism utilizes an offline bank of successful trajectories to retrieve actionable insights via task similarity, offering robust guidance in a training-free manner.

\section{Implementation Details}
\label{sec:implementation}
\textbf{Benchmarks and Datasets.} 
We evaluate our framework on two primary benchmarks: \textbf{WebVoyager} and \textbf{OnlineMind2Web}. Due to regional access and verification restrictions, we select a representative subset of \textbf{12 websites} from WebVoyager for evaluation. For OnlineMind2Web, which involves real-time web interaction, we successfully conduct tasks across \textbf{291 websites}, discarding 9 tasks where the target URLs were inaccessible at the time of testing. These datasets provide a rigorous testbed for long-horizon reasoning and cross-domain UI navigation.

\textbf{Agent Backbones.} 
To demonstrate the model-agnostic nature of our dual-memory approach, we employ three multimodal models as agents: \textbf{Claude-3.7-Sonnet}, \textbf{Claude-Sonnet-4}, and the open-source \textbf{Qwen3-VL-32B}. All models are tasked with operating web pages and executing user instructions within a training-free prompting framework.

\textbf{Infrastructure and Environment.} 
Experiments are conducted on a high-concurrency platform built on \textbf{Kubernetes (K8s)}. We deploy a cluster of \textbf{64 concurrent nodes}, each hosting a virtual computer environment. The agents interact with these environments via \texttt{kubectl} to perform browser actions and capture visual feedback. To balance task complexity and computational cost, we limit each task to a maximum of \textbf{60 steps}. 

\textbf{Dual-Memory Configuration.} 
For the internal memory, while the baseline maintains a sliding window of the last $k=5$ screenshots, our framework replaces this with a summary chain $\mathcal{M}^{int}$, retaining only the current observation $O_t$. The external memory is powered by an \textit{Insight Bank} distilled from \textbf{55k successful trajectories}. We utilize the \texttt{all-MiniLM-L6-v2} Sentence Transformer to calculate \textbf{cosine similarity} between the current query $Q$ and historical trajectories, retrieving the \textbf{Top-5} most relevant insights as strategic hints.

\textbf{Evaluation Protocol.} 
Following the interaction, task success is determined by a \textbf{GPT-4o judge}. The judge evaluates the model's final response and the \textbf{last five screenshots} of the trajectory to verify goal completion. This visual-textual evidence ensures a robust and objective assessment of the agent's performance in real-world navigation scenarios.

\begin{figure}[h]
    \centering
    \includegraphics[width=1.0\linewidth]{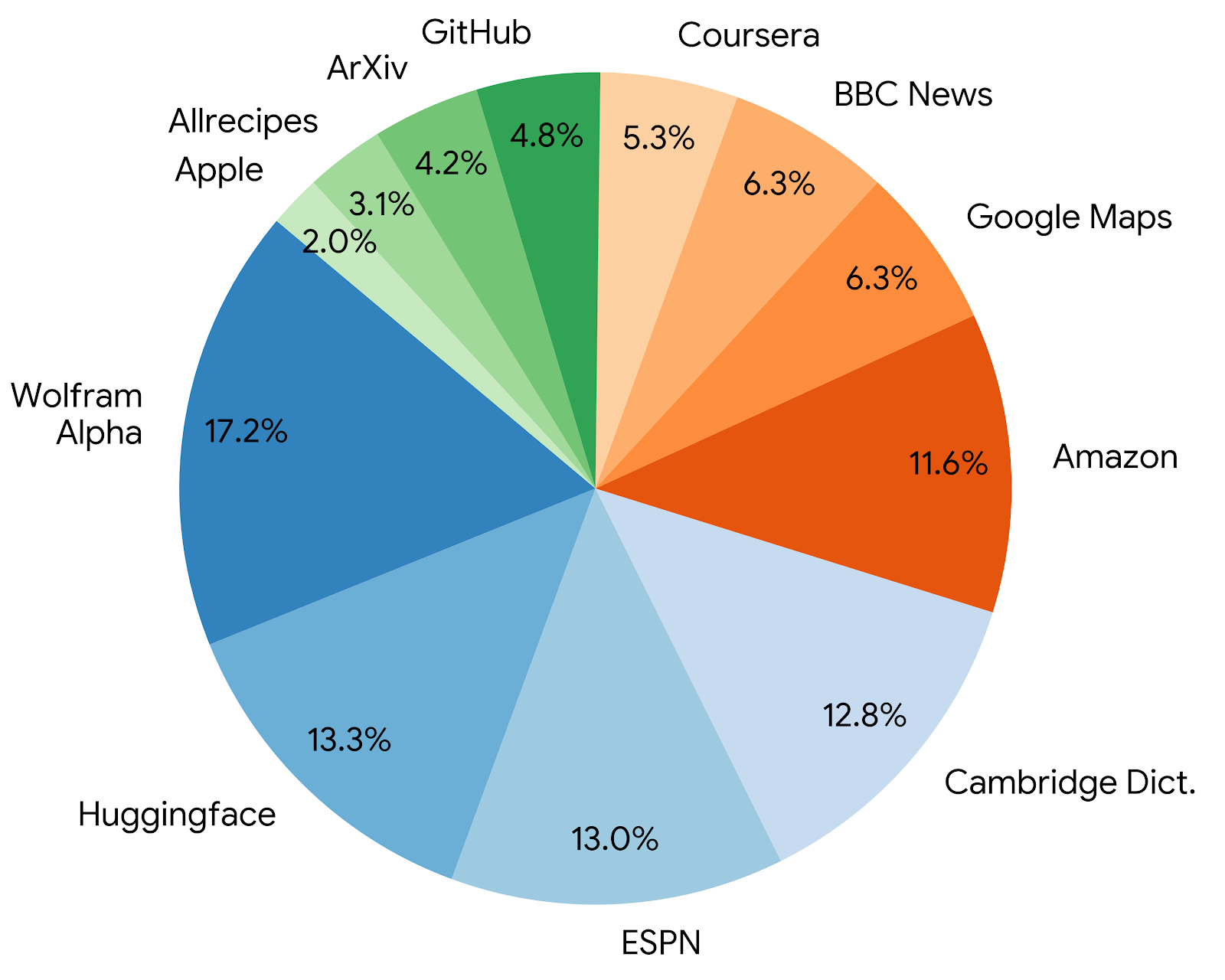} % Replace with your actual filename
    \caption{Distribution of the 55k trajectory data across 12 distinct web domains in the Insight Bank.}
    \label{fig:distribution}
    % \vspace{-0.5cm}
\end{figure}

\begin{figure}[h]
    \centering
    \includegraphics[width=1.0\linewidth]{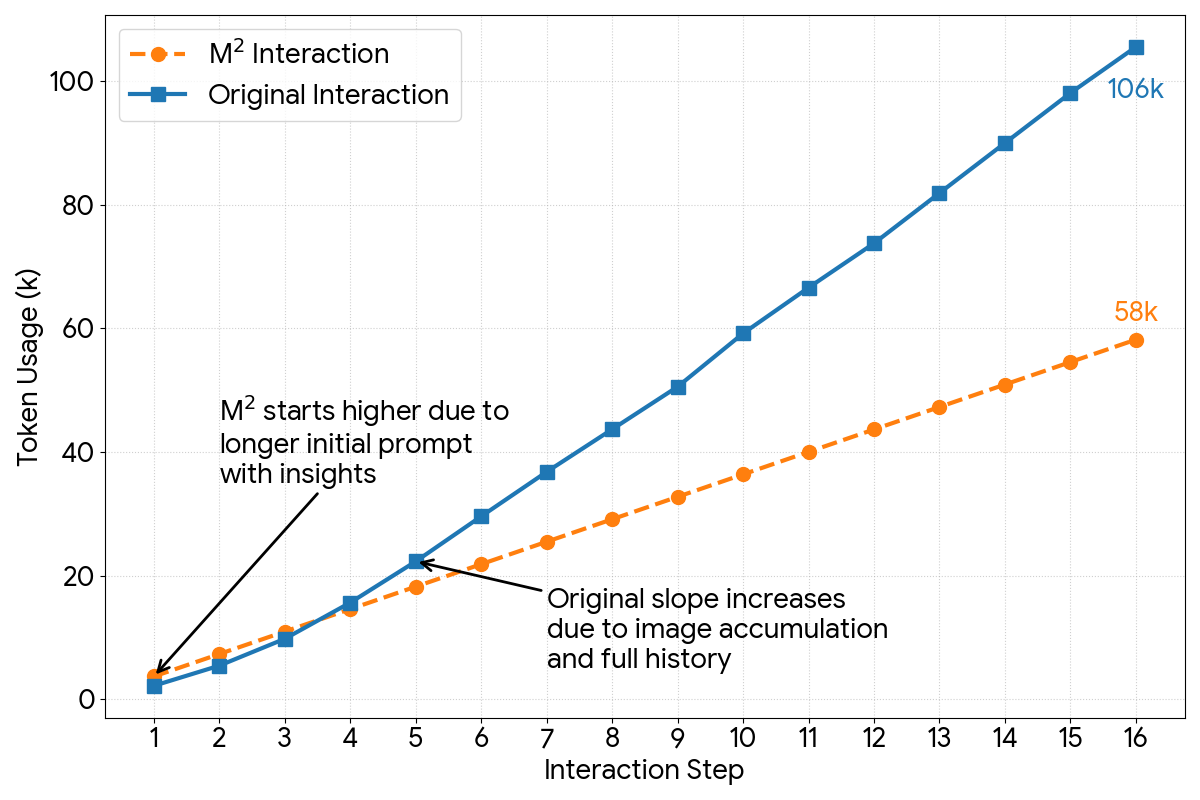} % Ensure filename matches
    \caption{Cumulative token consumption comparison on an Apple.com navigation task (16 steps). The agent uses Qwen3-VL-32B.}
    \label{fig:token_growth}
    % \vspace{-0.5cm}
\end{figure}

\section{Insight Bank Statistics \& Details}
\label{sec:insight_bank}
To ensure the generalization capability and robustness of our external memory, we construct a large-scale Insight Bank derived from approximately 55k successful execution trajectories. These trajectories span 12 diverse web domains, covering categories such as e-commerce, academic research, coding platforms, and utility tools. The user queries for all these trajectories are generated by LLM and are strictly distinct from the benchmarks used in the experiments.

As illustrated in Fig.~\ref{fig:distribution}, the data distribution is strategically balanced to prevent overfitting to specific website layouts. 
% The dataset is led by computational and technical domains, with \textbf{Wolfram Alpha} (17.2\%) and \textbf{Huggingface} (13.3\%) comprising a significant portion, followed closely by information-heavy sites like \textbf{ESPN} (13.0\%) and \textbf{Cambridge Dictionary} (12.8\%). E-commerce interactions are represented by \textbf{Amazon} (11.6\%) and \textbf{Apple} (2.0\%), while functional tools like \textbf{Google Maps} (6.3\%) and \textbf{GitHub} (4.8\%) ensure the agent learns to handle complex interactive elements. 
This diversity ensures that the extracted insights cover a wide spectrum of interaction logic, from static information retrieval to dynamic form manipulation.

Beyond the quantitative scale, qualitative analysis of stored insights is critical for effective retrieval augmentation. Unlike raw trajectory logs which record specific interactions, our Insight Bank stores generalized, high-leverage interaction rules. To illustrate this, consider a complex task on \textit{Apple.com} requiring the configuration of a 16-inch MacBook Pro with specific specifications (M4 Max chip, 32GB memory, etc.). Instead of memorizing the specific click sequence for this singular product, the extracted insights provide universal UI strategies applicable to similar shopping scenarios:

\begin{itemize}
    \item \textbf{Search Strategy:} When the exact query yields no results, strip the query to the core noun and use the sidebar model filter to drill down manually.

    \item \textbf{Interaction Order:} Always apply the ``Sort by Date/Price'' \textit{before} selecting specific filters. On many dynamic sites, changing the sort order triggers a page refresh that inadvertently resets active filters, causing the agent to lose progress.

    \item \textbf{State Validation:} After clicking ``Add to Cart/Bag,'' do not proceed immediately; verify the ``Cart Icon'' badge number has incremented. If it hasn't changed within 3 seconds, the click was likely intercepted by an overlay—trigger the click again.

    \item \textbf{Navigation Efficiency:} If a page fails to load or display properly, use the browser's ``Back" key (Alt+Left) to return to the previous page and resume the configuration process, rather than refreshing the page repeatedly.
\end{itemize}

By retrieving and injecting such structured, logic-driven insights into the prompt, our method enables the agent to reference explicit operational goals and avoid recurring navigation traps during inference.

\section{Token Consumption Growth Curve}
\label{sec:token_analysis}
To rigorously quantify the efficiency gains of our proposed framework, we conduct a step-by-step token consumption analysis on a representative task by Qwen3-VL-32B. The task required the agent to navigate \textit{Apple.com} to retrieve specific technical specifications and software compatibility information: \textit{``Find information on Apple website, and tell me the device weight of Apple Vision Pro and list 5 Built-in Apps it supports.''}

Fig.~\ref{fig:token_growth} illustrates the cumulative token usage for both the \textbf{Original} Interaction mode (full-history concatenation) and our \textbf{M$^2$} Interaction mode (dual-memory augmentation) over a 16-step trajectory.

The comparison reveals three distinct phases of token consumption dynamics:

\paragraph{1. Initial Overhead (Step 1):} 
As observed in the initial phase, the M$^2$ framework incurs a higher starting cost compared to the baseline. Specifically, at Step 1, M$^2$ consumes approximately $\sim$3.6k tokens, exceeding the Original method by $\sim$1.7k tokens. This overhead is intentional and attributable to the \textit{Insight Retrieval Augmentation} module, which injects retrieved guidelines and defensive instructions into the system prompt before the first action is taken. This upfront investment is designed to ``prime" the agent with expert knowledge.

\paragraph{2. The Crossover Point (Steps 3--5):} 
The efficiency advantage of the baseline is short-lived. The curves intersect between Step 3 and Step 4. By Step 4, the cumulative cost of the Original method accelerates rapidly as it begins to stack raw history (screenshots and entire text) from previous steps. In contrast, M$^2$ employs \textit{Dynamic Trajectory Summarization}, discarding raw observations in favor of concise textual state updates. Consequently, the consumption of the original method surpasses M$^2$ almost immediately after the several interaction loop begins.

\paragraph{3. Long-Term Growth Trends (Step 5--16):} 
The divergence becomes drastic as the trajectory lengthens. 
\begin{itemize}
    \item \textbf{Original Interaction (Rapid Growth):} The baseline exhibits a super-linear growth trend. By Step 16, the context window is burdened with full interaction history, driving the total consumption to \textbf{106k tokens}. This exponential bloat not only increases cost but also introduces significant visual noise.
    \item \textbf{M$^2$ Interaction (Gradual Growth):} In contrast, our method maintains a stable growth rate. By condensing history into summaries, the marginal cost per step remains constant ($\sim$3.7k tokens). By Step 16, the total consumption is only \textbf{58k tokens}, representing a \textbf{45.3\% reduction} compared to the baseline.
\end{itemize}

This analysis confirms that while M$^2$ requires a minimal cold-start ``token investment'' to prime the agent with expert knowledge, it effectively solves the context explosion problem, making it a highly scalable solution for long-horizon web navigation.

\begin{figure*}[t]
    \centering
    \includegraphics[width=\textwidth]{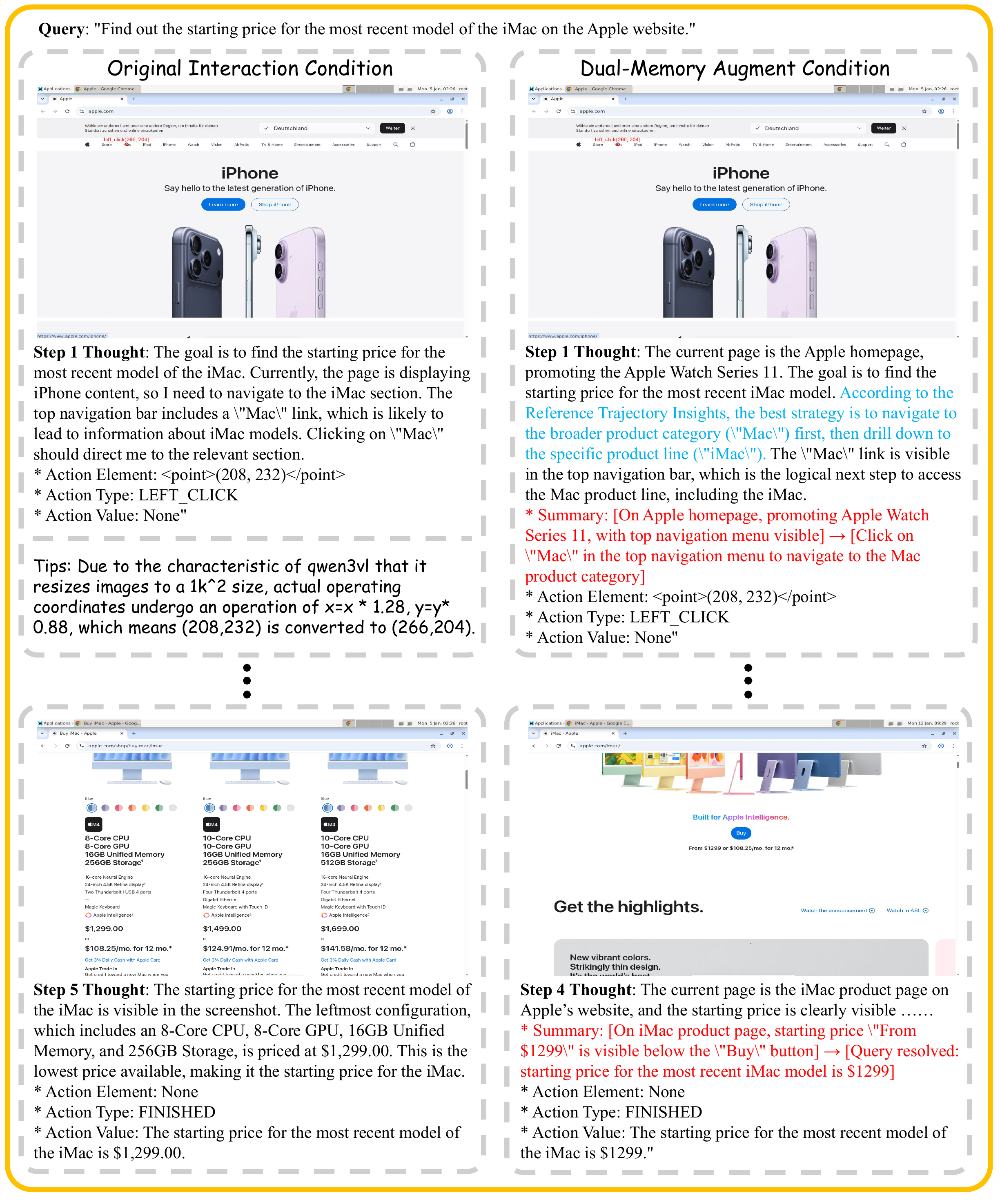}
    \caption{Comparison of agent responses between the original interaction form and the M$^2$ interaction form.}
    \label{fig:case_study}
\end{figure*}

\begin{figure*}[t]
    \centering
    \includegraphics[width=\textwidth]{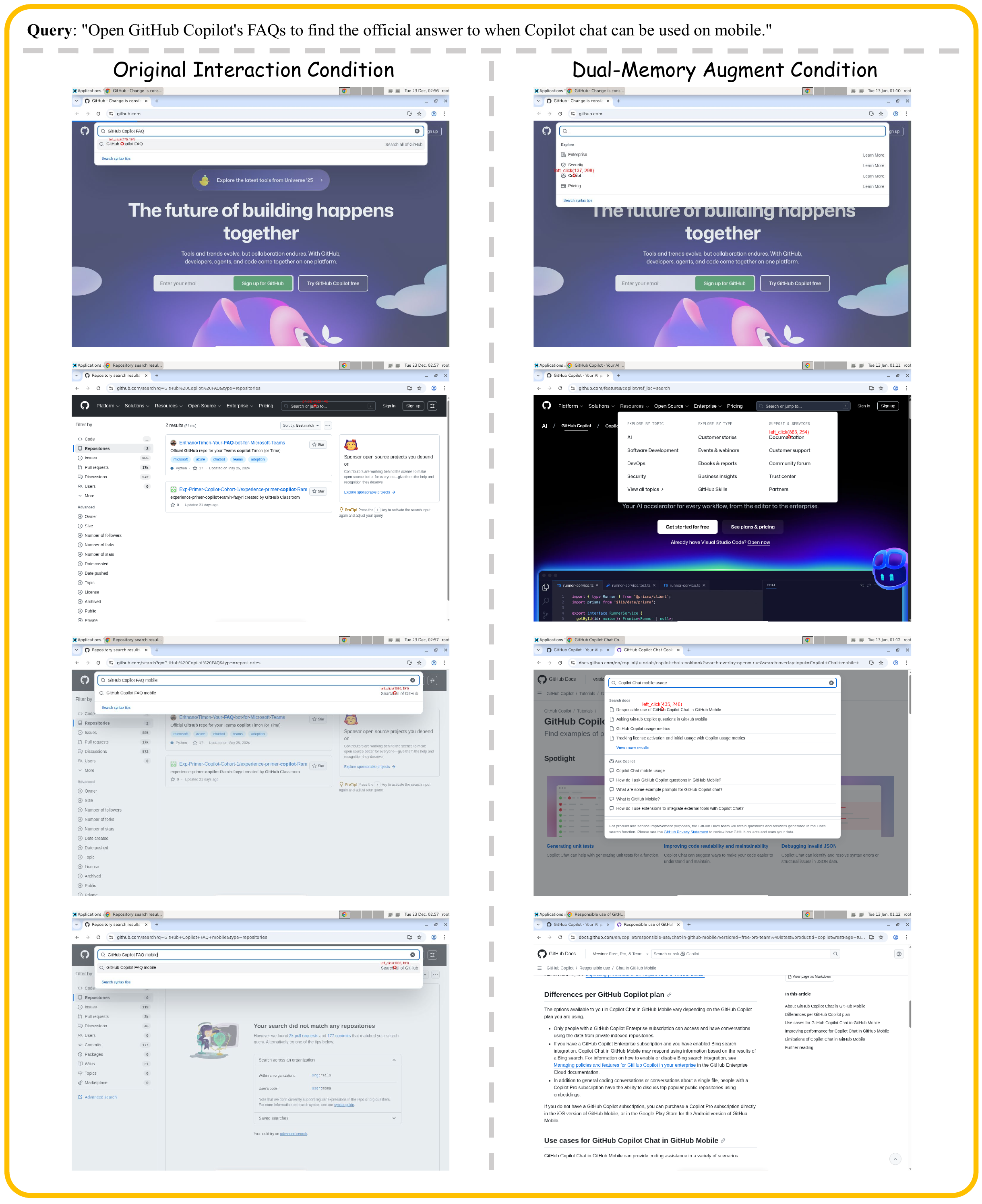}
    \caption{Qualitative comparison on the GitHub task. \textbf{Left (Original):} Agent gets trapped in a loop of irrelevant repository results due to context redundancy. \textbf{Right (M$^2$):} Agent navigates hierarchically to the documentation page and uses a precise search, maintaining a clear task progress via memory summarization.}
    \label{fig:compare_case_study}
\end{figure*}

\begin{figure*}[t]
    \centering
    \includegraphics[width=\textwidth]{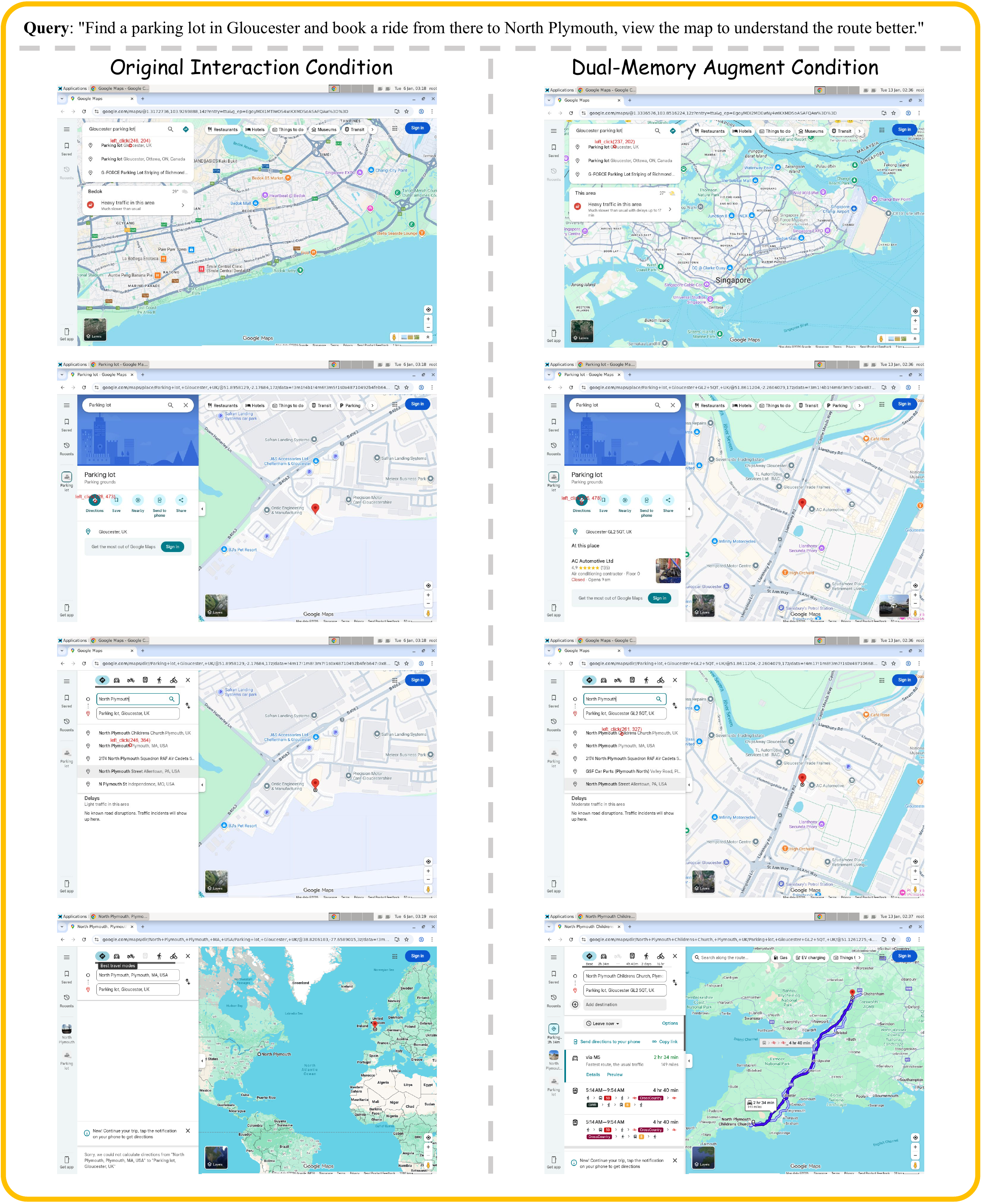}
    \caption{Qualitative comparison on a geographic disambiguation task. \textbf{Left (Original):} The agent suffers from context hallucination, ignoring the established UK starting point and selecting a US destination (North Plymouth, MA), resulting in a routing failure. \textbf{Right (M$^2$):} Guided by the external insight to prioritize the first dropdown suggestion and maintaining internal context consistency, the agent correctly identifies the local UK destination.}
    \label{fig:compare_case_study_2}
\end{figure*}

\section{Qualitative Analysis}
\label{sec:qual_analysis}
\subsection{Comparison of Response Structure Visualization}
Fig.~\ref{fig:case_study} illustrates the structural differences in agent responses between the baseline and our proposed method.

\textbf{Original Interaction Condition (Left):}
The baseline agent follows a standard ReAct-style output format. The response consists solely of a \texttt{Thought} block, where the agent reasons based on the immediate visual observation, followed directly by the \texttt{Action} parameters. The context relies implicitly on the raw history of previous turns.

\textbf{Dual-Memory Augment Condition (Right):}
In contrast, the M$^2$ framework introduces two distinct structural enhancements to the response format:
\begin{itemize}
    \item \textbf{Insight-Aware Reasoning:} Within the \texttt{Thought} block, the agent explicitly references the external memory (e.g., \textit{``According to the Reference Trajectory Insights...''}), demonstrating that high-level strategic guidance is actively incorporated into the planning process before action selection.
    \item \textbf{Structured Summarization:} A unique \texttt{Summary} field is generated alongside the action. This field adheres to a strict template (\texttt{[Brief current page state] $\to$ [Brief action taken]}), which serves as the compressed internal memory update for the subsequent step, replacing the need to store raw observation logs.
\end{itemize}

\subsection{Case Study: GitHub Copilot FAQ Retrieval}

We analyze a retrieval task: \textit{``Open GitHub Copilot's FAQs to find the official answer to when Copilot chat can be used on mobile.''} (Fig.~\ref{fig:compare_case_study}).

\noindent\textbf{Baseline Failure (Context Overload):} 
% The baseline agent falls into a ``global search trap,'' repeatedly querying the homepage and retrieving irrelevant repositories instead of documentation. The accumulation of visually redundant search pages in the full context window obscures state awareness, causing the agent to loop continuously without attempting a different navigation strategy.
The baseline agent falls into a ``global search trap,'' repeatedly querying the homepage. Specifically, the agent consistently triggers the ``Search all of GitHub'' action and receives a static ``No results found'' page; interpreting this visual invariance as a system non-response, the agent mistakenly retries the click indefinitely, obscuring state awareness in the context window.

\noindent\textbf{M$^2$ Success (Insight \& Summary-Driven Navigation):} 
In contrast, M$^2$ bypasses the noisy global search in favor of hierarchical navigation (\texttt{Copilot} $\to$ \texttt{Resources} $\to$ \texttt{Documentation}). This is enabled by M$^2$, which maintains a clean execution log and professional guidance. By filtering out visual noise, the concise context allows the agent to track its progress accurately and eventually execute a precise local search within the documentation to retrieve the answer.

\subsection{Case Study: Disambiguating Geographic Locations}

We examine a navigation task susceptible to geographic ambiguity: \textit{``Find a parking lot in Gloucester and book a ride from there to North Plymouth, view the map to understand the route better.''} (Fig.~\ref{fig:compare_case_study_2}).

\noindent\textbf{Baseline Failure (Context Hallucination):}
The baseline agent initially identifies the starting point in \textit{Gloucester, UK}. However, when searching for the destination ``North Plymouth,'' it suffers from a \textit{contextual hallucination}. Ignoring the established UK context, it selects the second suggestion, ``North Plymouth, MA, USA.'' This results in an infeasible trans-Atlantic routing error.

\noindent\textbf{M$^2$ Success (Insight \& Context Synergy):}
The M$^2$ agent successfully resolves this ambiguity through two synergistic mechanisms:
\begin{itemize}
    \item \textbf{Insight-Driven Selection Strategy (External Memory):} Guided by the specific insight \textit{``Preferably select the first suggestion from the dropdown list to ensure accuracy,''} the agent prioritizes the top-ranked suggestion provided by the search engine (which naturally prioritizes proximity to the current map view/start point).
    \item \textbf{Contextual Consistency (Internal Memory):} The agent's reasoning explicitly validates the choice: \textit{``matches the intended destination in the UK (as Gloucester is in the UK).''} By adhering to the specific context rather than relying on intrinsic bias, it correctly selects ``North Plymouth, UK,'' enabling a valid route calculation.
\end{itemize}

\begin{figure*}[t]
    \centering
    \includegraphics[width=\textwidth]{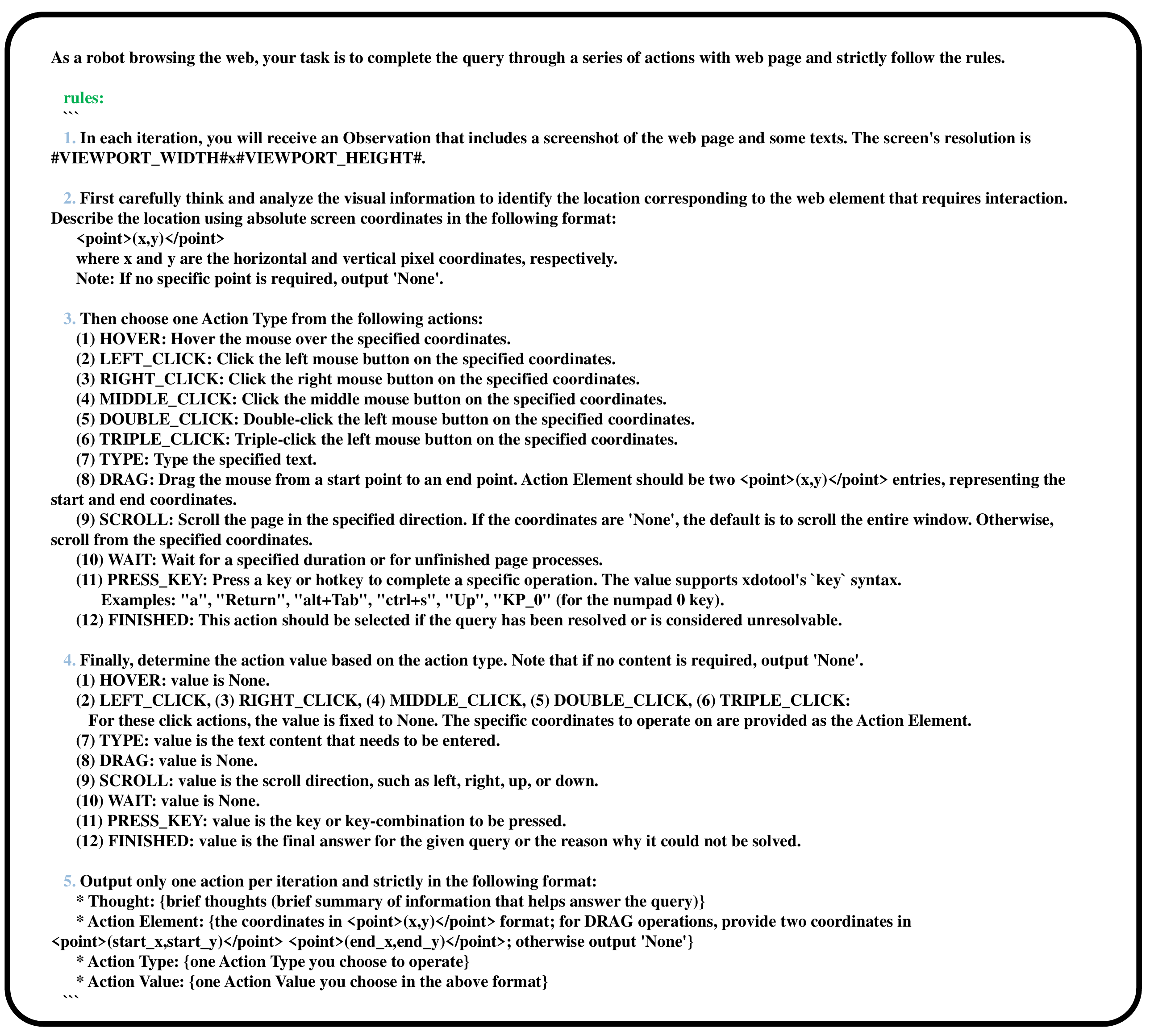}
    \caption{Original system prompt used by the agent.}
    \label{fig:system_prompt}
\end{figure*}

\begin{figure*}[t]
    \centering
    \includegraphics[width=\textwidth]{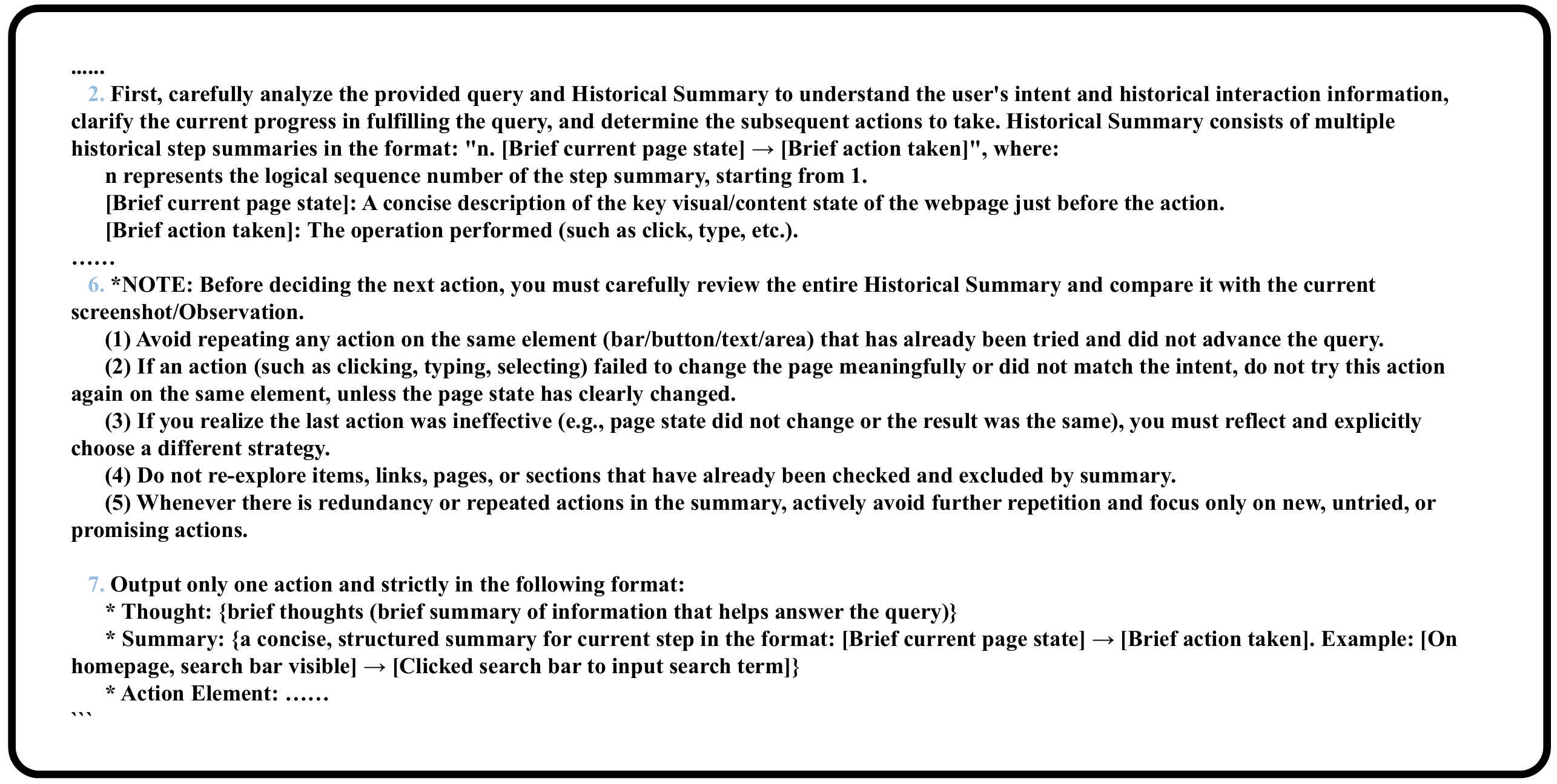}
    \caption{Update of Dynamic Trajectory Summarization compared to the original system prompt.}
    \label{fig:in_sys_prompt}
\end{figure*}

\begin{figure*}[t]
    \centering
    \includegraphics[width=\textwidth]{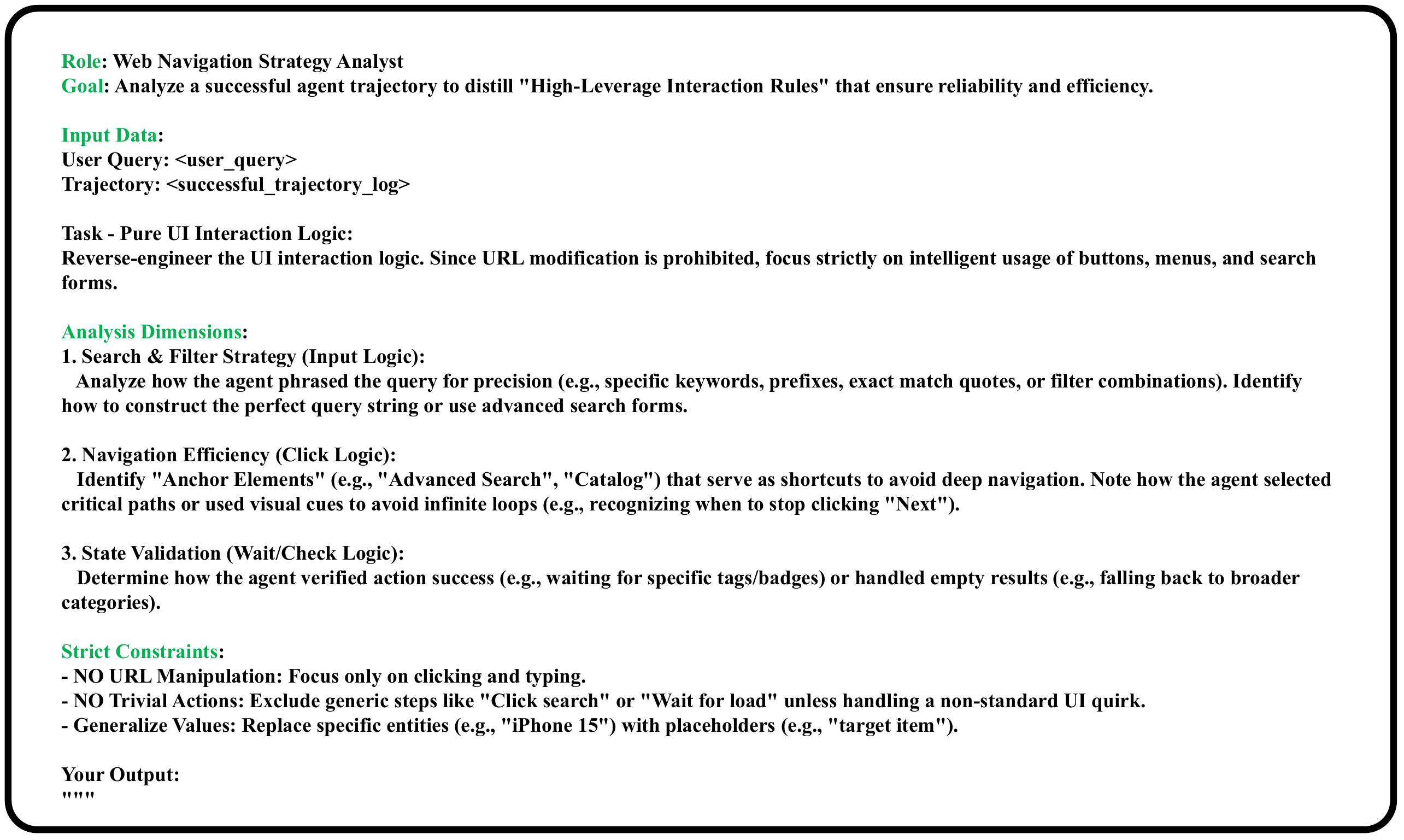}
    \caption{Prompt $P_{ext}$ used by Abstractor to extract insight.}
    \label{fig:abstractor_prompt}
\end{figure*}

\begin{figure*}[t]
    \centering
    \includegraphics[width=\textwidth]{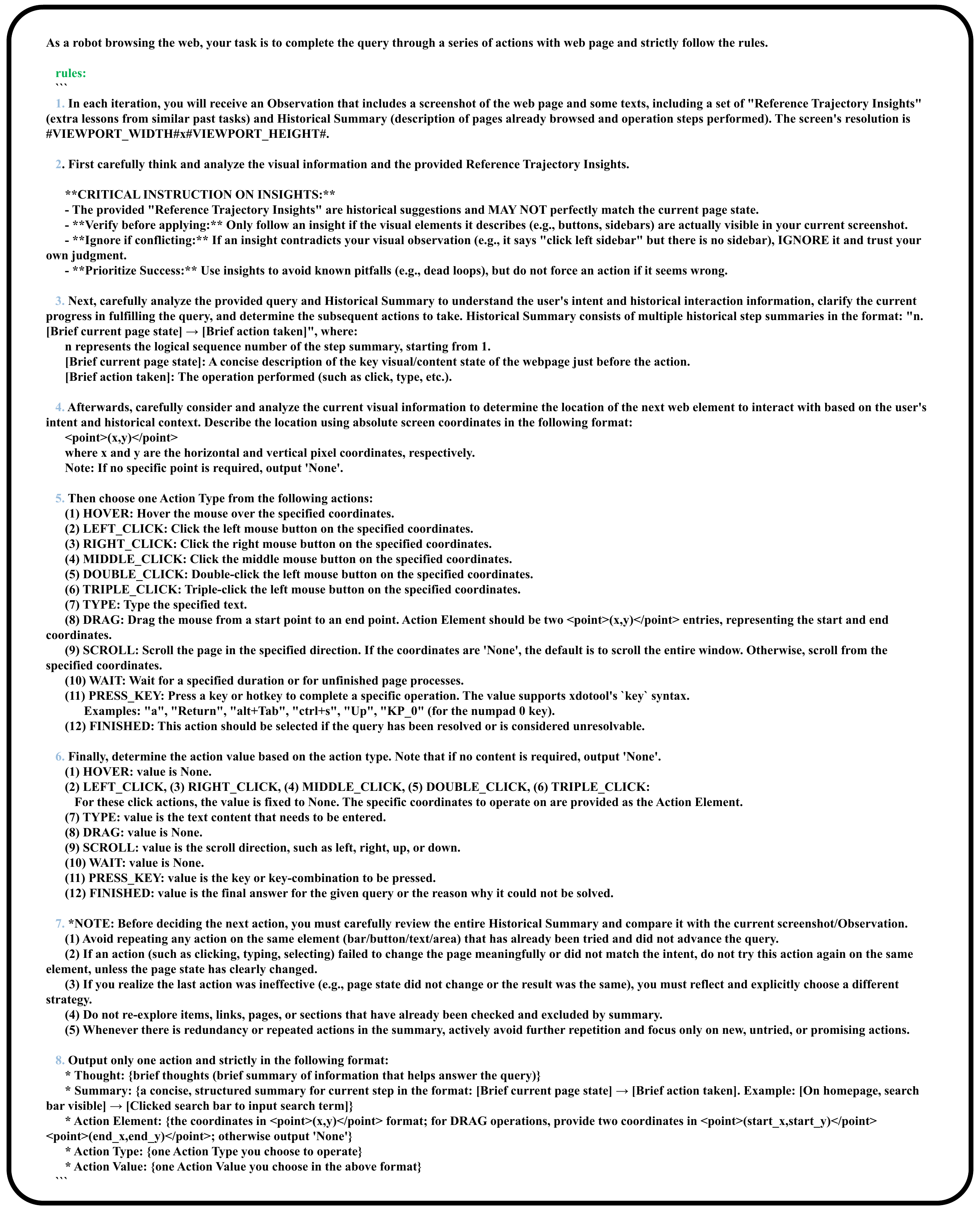}
    \caption{System prompts $P^{\prime}_{sys}$ used in the final M$^2$ framework.}
    \label{fig:final_system_prompt}
\end{figure*}

\section{Prompt Templates}
\label{sec:prompt_template}
\subsection{Original System Prompt $P_{sys}$ and Action Space}
As shown in Fig.~\ref{fig:system_prompt}, the system prompt $P_{sys}$ establishes model's role as an autonomous web browsing agent tasked with resolving user queries through sequential interaction with web interfaces. At each step, the agent receives a multimodal observation consisting of a webpage screenshot and semantic text, grounded by the specific viewport resolution. 
% To bridge visual perception and action execution, the agent must analyze the layout to identify target elements and represent them using absolute screen coordinates. 
The action space is designed to simulate human interactions, encompassing mouse operations (e.g., \texttt{HOVER}, \texttt{LEFT\_CLICK}, \texttt{DRAG}), keyboard inputs (\texttt{TYPE}, \texttt{PRESS\_KEY}), and navigational controls (\texttt{SCROLL}, \texttt{WAIT}, and page transitions). To ensure execution, the model is constrained to generate a strictly formatted four-part output: a \textit{Thought} summarizing the reasoning process, an \textit{Action Element} specifying the target coordinates (e.g., \texttt{<point>(x,y)</point>}), an \textit{Action Type} selected from the predefined set, and an \textit{Action Value} containing necessary parameters such as input text or scroll direction.

\subsection{$P^{in}_{sys}$ and $s_t$ for Dynamic Trajectory Summarization}
To accommodate the integration of the dynamic memory mechanism, we refine the standard system instructions into a history-aware prompt, denoted as $P^{in}_{sys}$ (Fig.~\ref{fig:in_sys_prompt}). Unlike the baseline which relies solely on immediate observations, $P^{in}_{sys}$ explicitly conditions the agent's reasoning on a ``Historical Summary''---a compressed chronological log of prior interactions ($\mathcal{M}_t^{int}$). This summary enforces a structured format where each entry $s_t$ encapsulates the pre-action state and the operation performed, exemplified by: $s_t =$ \texttt{[On Apple Watch Series 11 landing page with "Buy" button visible] $\to$ [Click the "Buy" button to access pricing details]}. Furthermore, $P^{in}_{sys}$ incorporates strict negative constraints (Rule 6) that compel the agent to cross-reference the current observation against the historical trace to actively prune redundant explorations and prevent execution loops. Finally, the output space is augmented to require the generation of the current step's summary alongside the action, ensuring the memory stream is autonomously maintained.

\subsection{$P_{ext}$ for Insight Retrieval Augmentation}
The insight extraction prompt $P_{ext}$ (Fig.~\ref{fig:abstractor_prompt}) configures the Model to act as a ``Web Navigation Strategy Abstractor,'' tasked with distilling generalizable ``High-Leverage Interaction Rules'' from historical successful trajectories. Unlike standard summarization, this prompt enforces a \textit{Pure UI Interaction} constraint, strictly prohibiting URL manipulation to ensure the extracted insights focus on intelligent interface usage---such as identifying ``Anchor Elements'' for navigation shortcuts or constructing precise search queries. The analysis is stratified into three specific dimensions: \textit{Search \& Filter Strategy} (optimizing input logic), \textit{Navigation Efficiency} (optimizing click paths), and \textit{State Validation} (visual logic for error avoidance). To ensure transferability, the prompt mandates the generalization of specific values and the exclusion of trivial actions. The output is constrained to a strict, machine-parsable text format, consisting of newline-separated insights prefixed with topic tags (e.g., \texttt{[Search Strategy]}), enabling seamless storage and retrieval within the Insight Bank.

\subsection{$P^{\prime}_{sys}$ applicable to M$^2$}
Compared to the initial baseline prompt, the final version $P^{\prime}_{sys}$ (Fig.~\ref{fig:final_system_prompt}) incorporates substantial modifications to support the dual-memory architecture and ensure robust execution. First, the observation space (Rule 1) is augmented to include both ``Reference Trajectory Insights'' (External Memory) and a ``Historical Summary'' (Internal Memory), providing the agent with global strategic guidance and local state context. Second, a defensive verification mechanism is explicitly codified (Rule 2), instructing the agent to critically evaluate retrieved insights against the current visual screenshot and prioritize actual observations over historical suggestions to prevent context mismatch. Third, the reasoning process is constrained by history-aware negative constraints (Rule 7), effectively barring the agent from repeating ineffective actions or re-exploring visited states identified in the summary. Finally, the output space (Rule 8) is expanded to require the generation of a structured \texttt{Summary} for the current step, enabling the iterative maintenance of the internal memory stream without external intervention.


% You can have as much text here as you want. The main body must be at most $8$
% pages long. For the final version, one more page can be added. If you want, you
% can use an appendix like this one.

% The $\mathtt{\backslash onecolumn}$ command above can be kept in place if you
% prefer a one-column appendix, or can be removed if you prefer a two-column
% appendix.  Apart from this possible change, the style (font size, spacing,
% margins, page numbering, etc.) should be kept the same as the main body.
%%%%%%%%%%%%%%%%%%%%%%%%%%%%%%%%%%%%%%%%%%%%%%%%%%%%%%%%%%%%%%%%%%%%%%%%%%%%%%%
%%%%%%%%%%%%%%%%%%%%%%%%%%%%%%%%%%%%%%%%%%%%%%%%%%%%%%%%%%%%%%%%%%%%%%%%%%%%%%%

\end{document}